\documentclass[12pt,a4paper]{article}
\usepackage[utf8]{inputenc}
\usepackage{amsmath,amsfonts,amssymb}
\usepackage{graphicx}
\usepackage{booktabs}
\usepackage{array}
\usepackage{longtable}
\usepackage{multirow}
\usepackage{url}
\usepackage{natbib}
\usepackage{placeins}
\usepackage{listings}
\usepackage{geometry}
\usepackage{hyperref}
\hypersetup{
    colorlinks=true,
    citecolor=blue,
    linkcolor=black,
    urlcolor=blue,
    pdftitle={Automatic Prompt Generation via Adaptive Selection of Prompting Techniques}
}
\begin{document}

\title{Automatic Prompt Generation via Adaptive Selection of Prompting Techniques}

\author{
\textbf{Yohei Ikenoue, Hitomi Tashiro, Shigeru Kuroyanagi} \\
Spike Studio Inc. \\
\texttt{\{yohei.ikenoue, hitomi, shigeru\}@spikestudio.jp}
}

\date{}

\maketitle

\begin{abstract}
Prompt engineering is crucial for achieving reliable and effective outputs from large language models (LLMs), but its design requires specialized knowledge of prompting techniques and a deep understanding of target tasks. To address this challenge, we propose a novel method that adaptively selects task-appropriate prompting techniques based on users' abstract task descriptions and automatically generates high-quality prompts without relying on pre-existing templates or frameworks. The proposed method constructs a knowledge base that associates task clusters, characterized by semantic similarity across diverse tasks, with their corresponding prompting techniques. When users input task descriptions, the system assigns them to the most relevant task cluster and dynamically generates prompts by integrating techniques drawn from the knowledge base. An experimental evaluation of the proposed method on 23 tasks from BIG-Bench Extra Hard (BBEH) demonstrates superior performance compared with standard prompts and existing automatic prompt-generation tools, as measured by both arithmetic and harmonic mean scores. This research establishes a foundation for streamlining and standardizing prompt creation, enabling non-experts to effectively leverage LLMs.
\end{abstract}

\section{Introduction}

Large language models (LLMs) have significantly advanced natural language processing, achieving human-level performance on many benchmarks and surpassing humans on certain tasks. To achieve reliable and effective outputs from LLMs, the design of input instructions, known as prompts, is crucial. The systematic development of these prompts, referred to as prompt engineering, plays a central role in determining output quality.

However, current approaches to prompt engineering remain highly manual, dependent on expert knowledge, and reliant on extensive trial and error, which creates a substantial barrier to broader LLM adoption. For general users without expertise in prompt design, it is particularly challenging to translate abstract objectives into concrete prompts that yield effective results. This limitation creates demand for new approaches that allow users, regardless of their background, to leverage LLM capabilities through simple, non-technical input.

To address this challenge, we propose a prompt generation method that automatically applies task-appropriate prompting techniques based on user-provided task descriptions. The goal is to streamline and standardize prompt creation, enabling diverse users to exploit LLM capabilities through simple, high-level input. Specifically, the proposed method constructs a knowledge base that associates task clusters, characterized by semantic similarity across diverse tasks, with their corresponding prompting techniques. When users input task descriptions, the system assigns them to the most relevant task cluster and dynamically generates high-quality, task-adapted prompts by integrating techniques from the knowledge base, without relying on pre-existing templates or frameworks.

\section{Related Work}

As the importance of prompt engineering has grown, research in this field has advanced rapidly, with a strong focus on automation and optimization. These studies aim to support the discovery of effective prompts and facilitate broader use of LLMs.

A variety of prompting techniques have been proposed, including Chain-of-Thought (CoT) prompting for step-by-step reasoning \citep{wei2023chainofthoughtpromptingelicitsreasoning}, few-shot prompting for providing examples \citep{brown2020languagemodelsfewshotlearners}, and emotion prompting for incorporating affective instructions \citep{li2023largelanguagemodelsunderstand}. While such techniques have demonstrated effectiveness in specific problem settings, identifying which technique, or combination of techniques, is optimal for a given task still requires specialized knowledge and extensive trial and error. This limitation highlights the need for frameworks that can automatically select and apply prompting techniques.

Automatic Prompt Engineer (APE) \citep{zhou2023largelanguagemodelshumanlevel} is a prominent example of research on automatic prompt generation. APE discovers effective prompts by having LLMs generate and select instructions from task input–output examples, achieving notable results in prompt optimization for specific tasks. However, most existing automatic prompt generation methods, including APE, assume the availability of substantial input–output examples or seed prompts. In contrast, the method proposed in this paper generates task-specific prompts without relying on pre-existing templates, starting instead from abstract, high-level information such as users' objectives or tasks to be streamlined.

Other approaches attempt to generate prompts directly from natural language instructions. For example, Anthropic’s Prompt Generator \citep{anthropicpromptgenerator} allows users to describe task overviews and desired goals, then produces candidate prompts. This method similarly does not require seed prompts or I/O examples, making it an appropriate baseline for our evaluation. Accordingly, our experiments use Anthropic’s Prompt Generator as one of the comparison methods.

Building on this background, the novelty of this research lies in proposing a framework that generates effective prompts from scratch by: (1) analyzing and classifying the latent semantic structure of tasks from users' simple descriptions, (2) selecting from a pre-constructed knowledge base of diverse prompting techniques, and (3) dynamically combining techniques adapted to task characteristics. The aim is to realize a system that produces practical prompts from non-technical input, without requiring specialized expertise.

\section{Methodology}

\subsection{System Overview}

This section outlines the architecture and implementation of the proposed adaptive prompt generation system, which operates in two main phases: knowledge base construction and prompt generation. In the construction phase, diverse tasks are clustered by semantic similarity, and a knowledge base is created that links each cluster to effective prompting techniques. In the generation phase, user-provided task descriptions are analyzed, and the system dynamically produces high-quality, task-adapted prompts by selecting and integrating techniques from the knowledge base. The following subsections describe the system configuration and processing pipeline in detail.

\subsection{System Configuration}

This research employs multiple large language models (LLMs) in constructing the prompt generation system, assigning each model to roles aligned with its performance characteristics and processing requirements.

\begin{itemize}
\item \textbf{LLM (large language model): \texttt{gemini-2.5-pro}}
\item \textbf{Embedding Model: \texttt{gemini-embedding-exp-03-07}}
\end{itemize}

In subsequent sections, references to ``LLM'' or ``Embedding Model'' refer to these specific models.

\subsection{Knowledge Base Construction Phase}

This phase constructs the knowledge base required for prompt generation.

\begin{figure}[htbp]
\centering
\includegraphics[width=0.8\textwidth]{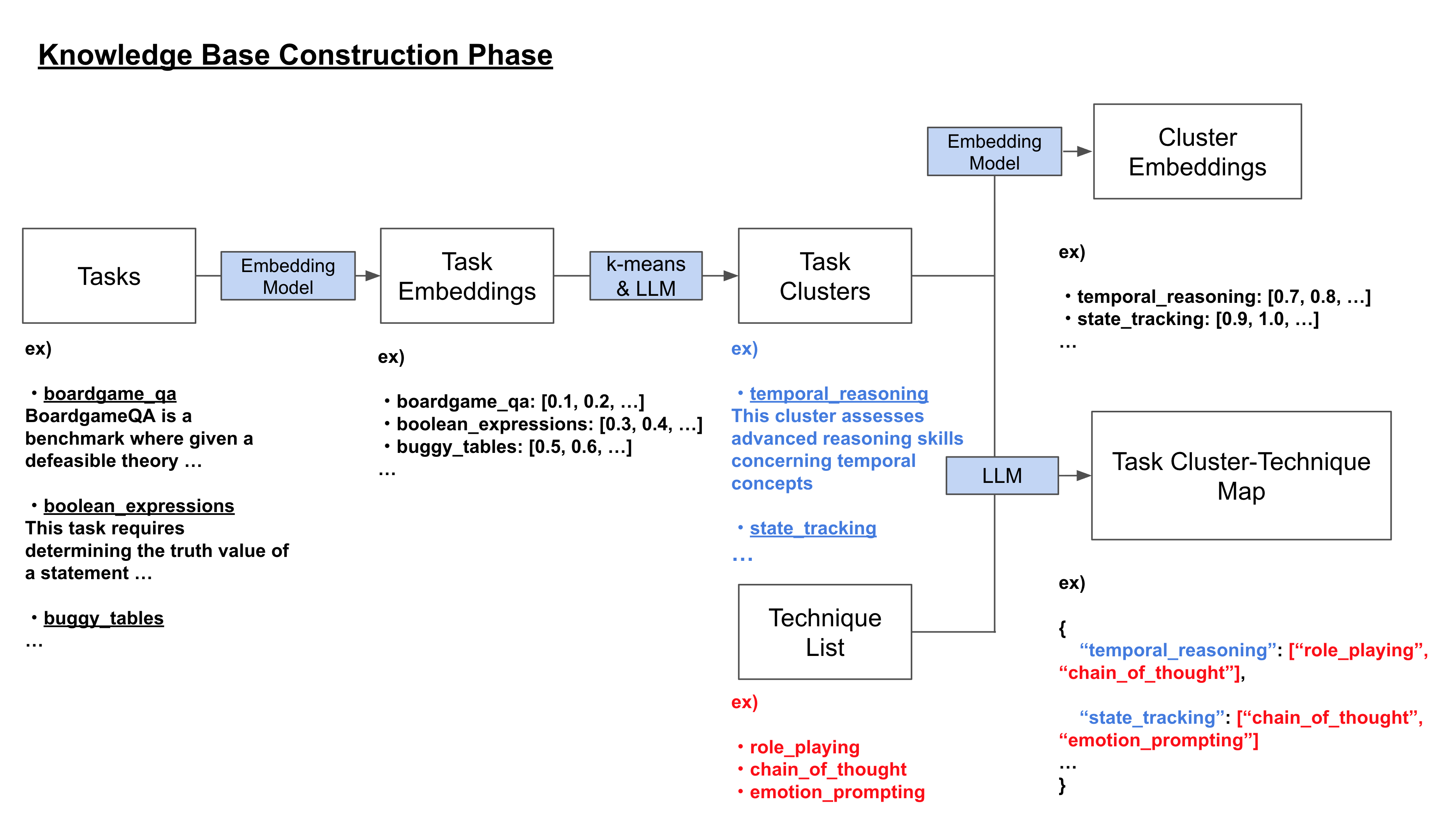}
\caption{Processing flow diagram of Knowledge Base Construction Phase}
\label{fig:knowledge_base_construction}
\end{figure}

\subsubsection{Task Cluster Definition and Semantic Space Construction}

The core of this method is the definition of task clusters that semantically classify and organize diverse user tasks. These clusters provide the foundation for selecting appropriate prompting techniques.

\textbf{Task Cluster Derivation Process:}
To ensure reproducibility, this study combines quantitative clustering with LLM-based semantic interpretation to derive task clusters.

\begin{itemize}
\item \textbf{Task Vectorization}: Each task’s name and description are combined into a text representation and converted into high-dimensional vectors using the Embedding Model. This embeds the semantic features of each task in vector space.
\item \textbf{Quantitative Determination of Optimal Cluster Number (K)}: The optimal number of clusters K is determined automatically using quantitative indicators. Specifically, k-means clustering \citep{MacQueen1967SomeMF} is executed across candidate values of K (ranging from 3 to fewer than the total number of tasks), and silhouette scores \citep{ROUSSEEUW198753} are computed for each. The clustering result with the highest silhouette score is adopted.
\item \textbf{LLM-based Cluster Semantic Assignment}: For each cluster identified, the LLM is provided with the list of tasks and generates a description of shared characteristics and required capabilities. This yields both detailed descriptions (\textit{cluster\_description}) and short identifiers (\textit{cluster\_id}) that express each cluster’s features in human-readable form.
\item \textbf{Task Cluster Semantic Space Construction}: The Embedding Model is then applied to each cluster’s detailed description (\textit{cluster\_description}) to map the clusters into high-dimensional vector space.
\end{itemize}

\subsubsection{Knowledge Base Construction for Task Cluster-Prompting Technique Mapping}

For each task cluster defined in Section 3.3.1, effective prompting techniques are mapped to construct a knowledge base. This process is designed to ensure reproducibility and logical consistency while maximizing the effectiveness of LLMs.

\clearpage

\textbf{Target Prompting Techniques:}
This study focuses on prompting techniques with high versatility and demonstrated effectiveness among those proposed in prior research on prompt engineering. Specifically, approximately 15 types of techniques are considered, as shown in Table \ref{tab:techniques}. These include methods that enhance prompts in various ways, such as improving instruction delivery to LLMs, guiding reasoning processes, and refining response quality.

\begin{table}[htbp]
\centering
\caption{Major Prompting Techniques Targeted (15 types)}
\label{tab:techniques}
\scriptsize
\begin{longtable}{p{3cm}p{4.5cm}p{8cm}}
\toprule
Category & Name & Description \\
\midrule
Role Assignment & Role Playing \citep{kong2024betterzeroshotreasoningroleplay} & Elicits domain-specific knowledge and reasoning patterns by assigning expert roles to the LLM. \\
\midrule
Emotional Stimulus & Emotion Prompting & Incorporates emotional cues into prompts to influence responses. \\
& Stress Prompting \citep{shen2025stresspromptdoesstressimpact} & Induces moderate stress conditions, grounded in psychological theory, to enhance LLM performance. \\
\midrule
Reasoning & Chain-of-Thought Prompting & Improves performance on complex reasoning tasks by requiring explicit step-by-step reasoning. \\
& Logic-of-Thought Prompting \citep{liu2025logicofthoughtinjectinglogiccontexts} & Strengthens logical reasoning by embedding propositional logic into the prompting process. \\
& Least-to-Most Prompting \citep{zhou2023leasttomostpromptingenablescomplex} & Decomposes complex problems into simpler sub-problems solved sequentially, using earlier answers to improve generalization. \\
& Thread of Thought Prompting \citep{zhou2023threadthoughtunravelingchaotic} & Enhances comprehension and generation by progressively summarizing and analyzing long, disorganized contexts. \\
& Plan-and-Solve Prompting \citep{wang2023planandsolvepromptingimprovingzeroshot} & Uses a two-stage process where LLMs first generate a plan and then execute it, improving reasoning accuracy. \\
& Skeleton-of-Thought Prompting \citep{ning2024skeletonofthoughtpromptingllmsefficient} & Reduces response latency by generating a response ``skeleton'' before completing details through parallel processing. \\
\midrule
Others & Decomposed Prompting \citep{khot2023decomposedpromptingmodularapproach} & Breaks down complex tasks into sub-tasks processed individually through prompts. \\
& Ignore Irrelevant Conditions \citep{wu2024instructinglargelanguagemodels} & Mitigates confusion by detecting and disregarding irrelevant information in problem statements. \\
& Highlighted Chain-of-Thought Prompting \citep{nguyen2025hothighlightedchainthought} & Improves accuracy and reduces hallucinations by highlighting essential information from long contexts before reasoning. \\
& Skills-in-Context Prompting \citep{chen2024skillsincontextpromptingunlockingcompositionality} & Enables LLMs to compose and apply basic skills in context, supporting generalization to novel problems. \\
& Automatic Information Filtering \citep{jiang2025enhancingrobustnesslargelanguage} & Preprocesses prompts by having LLMs identify and remove irrelevant information prior to reasoning. \\
& Scratchpad Prompting \citep{nye2021workscratchpadsintermediatecomputation} & Provides a ``draft space'' for intermediate reasoning steps, improving reliability in multi-step reasoning and calculations. \\
\bottomrule
\end{longtable}
\end{table}

\textbf{Knowledge Base Construction Process:}
The mapping of prompting techniques to each task cluster follows a set of clearly defined rules. The systematic procedure is as follows:

\begin{itemize}
\item \textbf{Input Information Presentation}: The LLM receives the cluster information (ID and detailed description) along with a list of all available prompting techniques (ID, name, description, and application cases).
\item \textbf{Constraint-based Technique Selection}: For each cluster, the LLM selects techniques according to four category constraints, resulting in three or four techniques per cluster.
\begin{itemize}
\item \textbf{Role Assignment (Fixed)}: Role playing is always included as a basic persona-setting technique applied to all clusters.
\item \textbf{Emotional Stimulus (Required)}: One technique must be chosen from Emotion Prompting or Stress Prompting. This category improves the LLM’s response style and motivation.
\item \textbf{Reasoning (Required)}: One reasoning technique must be chosen to promote complex thought processes and logical reasoning.
\item \textbf{Others (Optional)}: Zero or one additional technique may be selected to provide complementary support for specific problem characteristics.
\end{itemize}
\item \textbf{Knowledge Base Generation}: Finally, a Task Cluster–Prompting Technique Mapping is created, where each cluster ID is linked to a list of selected technique IDs (three to four techniques).
\end{itemize}

This approach systematically combines techniques with distinct roles—such as persona setting, emotional stimulation, logical reasoning, and complementary support—rather than merely listing potentially useful strategies. As a result, it establishes a multifaceted and robust foundation for prompt design in each cluster.

\subsection{Prompt Generation Phase}

This phase dynamically generates task-adapted prompts based on user input.

\begin{figure}[htbp]
\centering
\includegraphics[width=0.8\textwidth]{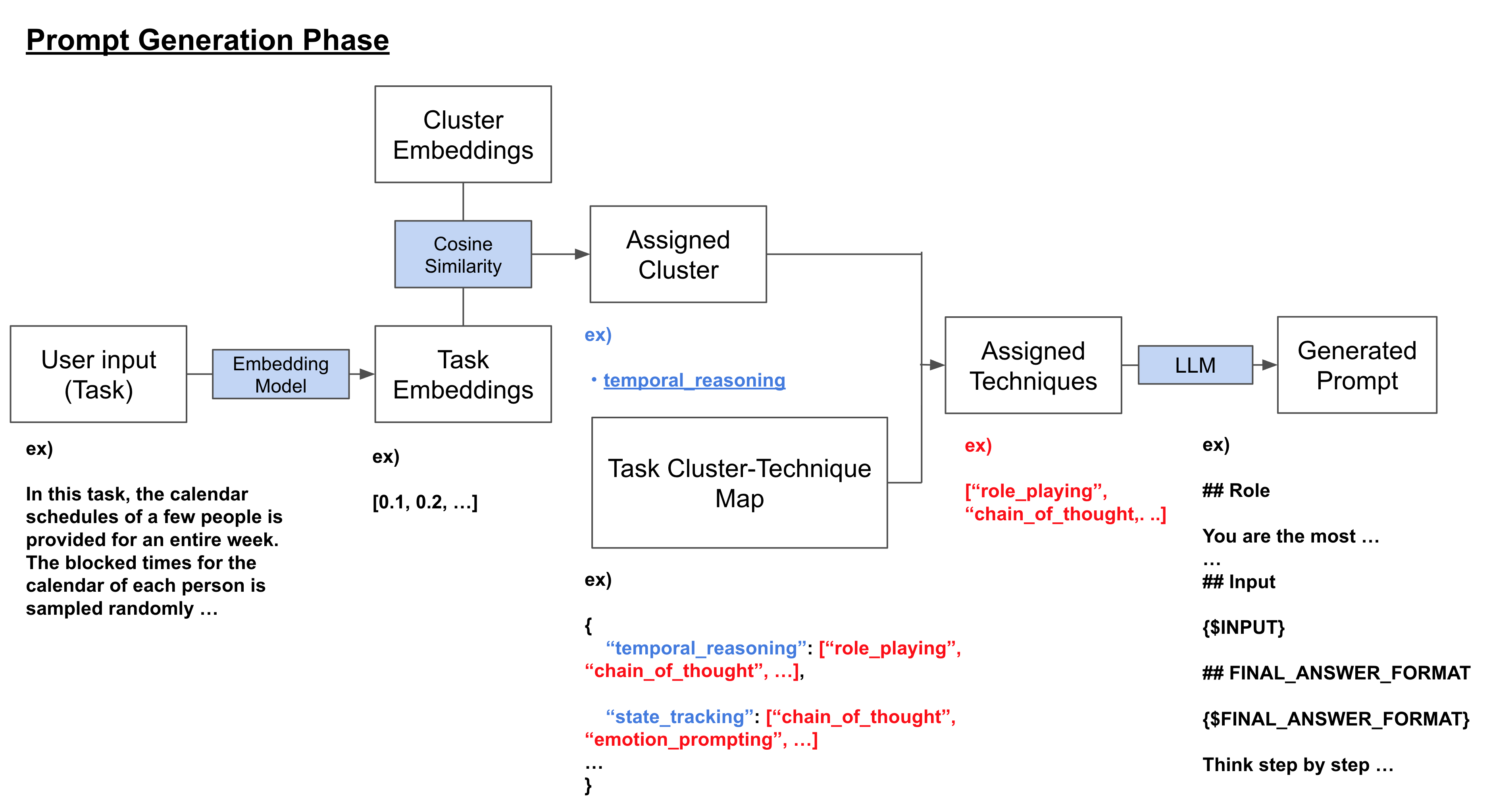}
\caption{Processing flow of the Prompt Generation Phase}
\label{fig:prompt_generation}
\end{figure}

\subsubsection{User Input Task Vectorization}
User-provided task descriptions and objectives in natural language are vectorized using the same Embedding Model employed for task cluster semantic space construction in Section 3.3.1. This process extracts the semantic features of each input task, enabling its positioning within the task cluster semantic space.

\subsubsection{User Input and Task Cluster Matching}
Cosine similarity \citep{salton_introduction_1983} is computed between the user input task vector from Section 3.4.1 and each task cluster vector stored in the database from Section 3.3.1. Based on these similarity scores, the task cluster most semantically aligned with the user’s input is identified. This step corresponds to interpreting the user’s implicit task intentions by associating them with one of the predefined clusters.

\subsubsection{Adaptive Selection of Prompting Techniques}
Using the task cluster identified in Section 3.4.2, the system references the Task Cluster–Prompting Technique Mapping Pool constructed in Section 3.3.2. From this pool, prompting techniques most suitable for the user’s task are selected. The techniques corresponding to the cluster with the highest similarity are adopted and supplied to the LLM as in-context guidance for generating the final prompt.

\subsubsection{Prompt Generation}
The user’s task description and the prompting techniques selected in Section 3.4.3 are provided to the LLM to generate prompts. This process produces high-quality prompts tailored to the user’s task intentions and characteristics, strategically designed to enhance LLM effectiveness.

\section{Experiments}

This section presents the experiments conducted to evaluate the effectiveness of the proposed method.

\subsection{Experimental Setup}

\subsubsection{Benchmark Dataset}

The evaluation was conducted using 23 tasks from BIG-Bench Extra Hard (BBEH) \citep{kazemi2025bigbenchextrahard}, an extension of the original BIG-Bench \citep{srivastava2023imitationgamequantifyingextrapolating}. BBEH includes tasks that require diverse reasoning capabilities and is widely adopted as an evaluation benchmark for LLMs.

BBEH updates BBH \citep{suzgun2022challengingbigbenchtaskschainofthought} by replacing several tasks with substantially more difficult variants, reflecting rapid advances in LLM capabilities. It comprises 23 reasoning tasks that span a broad range of cognitive abilities, including mathematical reasoning, language understanding, logical reasoning, and spatial cognition.

The rationale for selecting BBEH as the evaluation benchmark is two-fold:

\begin{enumerate}
\item \textbf{Clear measurement of prompt effects at high difficulty}: Because BBEH contains challenging tasks where standard prompts perform poorly, the impact of introducing prompting techniques can be assessed more distinctly.  
\item \textbf{Evaluation of versatility}: With 23 diverse tasks, BBEH enables a comprehensive assessment of general prompting techniques across multiple domains.  
\end{enumerate}

\subsubsection{Comparison Methods}

Two baseline methods were used as comparison targets for the proposed approach:

\begin{itemize}
\item \textbf{BBEH Original Prompts (Original paper)}: Performance using the prompts provided in the original BBEH paper for each task.  
\item \textbf{Anthropic Prompt Generator (Anthropic)}: Performance using prompts automatically generated by the Prompt Generator developed by Anthropic Inc. As discussed in the related work, this tool is an appropriate baseline because it also generates prompts directly from user instructions.  
\end{itemize}

To ensure statistical reliability, each method was evaluated over 10 independent trials. As evaluation metrics, we report average task accuracy across trials as well as the arithmetic and harmonic means aggregated across all tasks.

Representative examples of prompt templates for each method, including the proposed approach, are provided in Appendix A.

\textbf{Evaluation Metric Calculation Methods:}

\begin{itemize}
\item \textbf{Per-task arithmetic mean}: The arithmetic mean of accuracy rates across 10 trials for each task.  
\item \textbf{Arithmetic mean across tasks}: The arithmetic mean of per-task arithmetic means across all tasks.  
\item \textbf{Harmonic mean across tasks}: The harmonic mean of per-task arithmetic means across all tasks. To avoid division by zero, 1 is added to each per-task arithmetic mean before computing the harmonic mean. This calculation follows the procedure in the BIG-Bench Extra Hard proposal paper.  
\end{itemize}

\textbf{Model Configuration:}  
This experiment used \texttt{gemini-2.0-flash} for answer generation. \texttt{gemini-2.0-flash} is the standard evaluation model in BBEH benchmarks.  

\subsection{Experimental Procedure}  

The experimental procedure is standardized from prompt generation to correctness evaluation for both the proposed method and the comparison baseline (Anthropic's Prompt Generator), ensuring fair comparison.  

\subsubsection{Input Definition for Prompt Generators}  
Task information for each BBEH task is provided as input to both Anthropic's Prompt Generator and the proposed method. Specifically, the README files included in each BBEH task directory of the google-deepmind/bbeh GitHub repository \citep{bbeh2025} are used as instructional input. This ensures that both systems receive task overviews and objectives in a consistent format.  

\subsubsection{Output Prompt Standardization and Evaluation Logic}  
To enable consistent and automated evaluation using generated prompts, both Prompt Generators (Anthropic's and the proposed method's) are required to include two placeholders in their output:  

\begin{itemize}  
\item \verb|{$INPUT}|: Replaced at runtime with the problem statements from each BBEH task.  
\item \verb|{$FINAL_ANSWER_FORMAT}|: Replaced with instructions that strictly control the format of final answers. The embedded instruction string is identical to that used in the BBEH paper.  
\end{itemize}  

\begin{lstlisting}
Think step by step, and when you provide the final answer, please use the prefix "The answer is:" without any modification, and provide the answer directly, with no formatting, no bolding, and no markup. For instance: "The answer is: 42" or "The answer is: yes". If the question is multiple choice with a single correct answer, the final answer must only be the letter corresponding to the correct answer. For example, "The answer is: (a)".
\end{lstlisting}

\textbf{Evaluation Logic}: We use the official evaluate.py from the BBEH repository without modification. Specifically, final answers are extracted from LLM outputs using the following prefixes:

When an output contains any of these prefixes, it is treated as format-compliant, and the subsequent string is extracted and evaluated as the final answer. When none are present, the output is treated as format-noncompliant and excluded from accuracy scoring. This procedure yields a cleaner measure of each prompt’s ability to elicit correct answers.

\begin{lstlisting}
"The answer is:",
"The final answer is ",
"The final answer is: ",
"The answer is "
\end{lstlisting}

Details on format non-compliance tendencies between methods are provided in Appendix B1, which reports scores without applying this correction (i.e., treating non-compliant outputs as incorrect). Appendix B2 presents the proportion of format non-compliance occurrences for each method and task.

\textbf{Format Non-compliance Problem Examples}: Examples of format non-compliance include cases where output is interrupted before the final answer due to exceeding \texttt{gemini-2.0-flash}'s maximum output length (8192 tokens) during reasoning, as well as cases where the response deviates from the required format (e.g., \#\# ANSWER (A), ANSWER: (A), etc.).

\subsection{Experimental Results}

\subsubsection{Basic Experimental Results and Analysis}

\setcounter{table}{1}

\begin{table}[htbp]
\centering
\caption{Performance comparison of each prompt generation method on BBEH dataset (average across 10 trials)}
\label{tab:results}
\scriptsize
\begin{tabular}{lcccc}
\toprule
task name & Original & Anthropic & Ours & Ours (temperature-optimized) \\
\midrule
BoardgameQA & 42.1 & 42.8 & 43.6 & 46.0 \\
Boolean Expressions & 30.5 & 30.6 & 29.2 & 32.2 \\
Buggy Tables & 3.4 & 5.4 & 3.7 & 5.0 \\
Causal Understanding & 53.0 & 51.5 & 50.9 & 51.9 \\
DisambiguationQA & 49.6 & 45.5 & 47.4 & 41.0 \\
Dyck Languages & 12.1 & 13.8 & 23.9 & 20.9 \\
Geometric Shapes & 33.6 & 25.8 & 17.6 & 18.1 \\
Hyperbaton & 3.9 & 2.2 & 7.0 & 5.9 \\
Linguini & 14.2 & 14.5 & 13.9 & 14.0 \\
Movie Recommendation & 57.4 & 74.0 & 71.3 & 65.2 \\
Multistep Arithmetic & 10.8 & 17.4 & 22.5 & 20.8 \\
NYCC & 13.4 & 12.8 & 12.2 & 11.4 \\
Object Counting & 10.8 & 12.5 & 70.0 & 83.9 \\
Object Properties & 1.0 & 2.8 & 3.5 & 4.5 \\
SARC Triples & 34.8 & 30.6 & 34.4 & 35.3 \\
Shuffled Objects & 16.0 & 17.6 & 7.8 & 11.6 \\
Spatial Reasoning & 17.0 & 19.0 & 37.2 & 32.9 \\
SportQA & 21.1 & 25.0 & 23.0 & 24.5 \\
Temporal Sequences & 0.7 & 0.5 & 0.9 & 0.9 \\
Time Arithmetic & 42.8 & 41.0 & 39.4 & 41.6 \\
Web of Lies & 19.9 & 20.0 & 21.3 & 21.1 \\
Word Sorting & 24.5 & 20.6 & 21.7 & 22.1 \\
Zebra Puzzles & 37.8 & 42.7 & 40.7 & 45.7 \\
\midrule
\textbf{arithmetic mean across tasks} & \textbf{23.9} & \textbf{24.7} & \textbf{28.0} & \textbf{28.5} \\
\textbf{harmonic mean across tasks} & \textbf{9.7} & \textbf{10.5} & \textbf{12.5} & \textbf{13.3} \\
\bottomrule
\end{tabular}
\end{table}

The table above reports per-task arithmetic means over 10 trials for each method across 23 BBEH tasks. Aggregated scores are shown as the arithmetic mean across tasks and the harmonic mean across tasks. For harmonic mean calculation, 1 is added to the denominator to prevent division by zero, following the procedure in the BIG-Bench Extra Hard proposal paper.

The proposed method achieved an arithmetic mean of 28.0 across tasks, outperforming Original (23.9) and Anthropic (24.7). For the harmonic mean, the proposed method reached 12.5, again exceeding Original (9.7) and Anthropic (10.5).

\textbf{Analysis of arithmetic mean across tasks}: The arithmetic mean reflects average performance across all tasks. The proposed method’s score of 28.0 demonstrates consistently strong performance across BBEH’s diverse tasks. Its improvements over Original (+4.1) and Anthropic (+3.3) confirm its overall superiority.

\textbf{Analysis of harmonic mean across tasks}: The harmonic mean emphasizes low-scoring tasks. The proposed method’s score of 12.5 indicates stable performance even on the most difficult tasks. The gains over Original (+2.8) and Anthropic (+2.0) are larger than for the arithmetic mean, suggesting particular effectiveness in mitigating poor performance cases.

These findings indicate that the proposed method generates more effective prompts and better leverages LLM capabilities than both standard prompts and existing automatic generation methods in the high-difficulty BBEH benchmark. While notable improvements were observed in tasks such as Object Counting, Spatial Reasoning, and Zebra Puzzles, some tasks, including Boolean Expressions and Linguini, showed little or no improvement, indicating directions for future work.

\subsubsection{Per-task Analysis}

To analyze the proposed method’s performance in detail, we examine per-task accuracy improvements. As shown in Table \ref{tab:results}, the method produced varied effects across the 23 tasks, with outcomes differing significantly depending on task characteristics.

\textbf{Tasks Showing Improvement}: The proposed method improved performance in 15 tasks. Notable gains include Object Counting (+59.2), Spatial Reasoning (+20.2), Movie Recommendation (+13.9), Multistep Arithmetic (+11.7), and Dyck Languages (+11.8). Additional improvements were observed in Hyperbaton (+3.1), Zebra Puzzles (+2.9), Object Properties (+2.5), SportQA (+1.9), BoardgameQA (+1.5), Web of Lies (+1.4), Buggy Tables (+0.3), and Temporal Sequences (+0.2).

\textbf{Object Counting Success Analysis}: The Object Counting task showed the largest improvement. It requires extracting items meeting specific conditions from complex sentences and performing numerical calculations. Typical problems involve isolating possessions of a particular individual (e.g., ``me'') from distracting information (e.g., ``uncle,'' ``aunt'') and then computing totals through acquisition and loss histories (e.g., ``initial 22 → lose 20 → acquire 50 → lose 23 = 29''). The proposed method’s explicit step-tracking strategy aligns with this multi-stage process, yielding much higher accuracy than intuitive approaches.

\textbf{Tasks Showing Degradation}: Performance decreased in 8 tasks. The largest declines occurred in Geometric Shapes (-16.0) and Shuffled Objects (-8.2), with smaller drops in Time Arithmetic (-3.4), DisambiguationQA (-2.2), Causal Understanding (-2.1), Boolean Expressions (-1.3), NYCC (-1.2), and Linguini (-0.3).

\textbf{Geometric Shapes Failure Analysis}: The Geometric Shapes task requires recognizing figures from SVG path coordinates (e.g., ``M 0.0,9.0 L 23.94298,9.0'') and identifying the correct shape. Success depends on spatial cognition (the ability to mentally visualize shapes from coordinates). The language-focused reasoning strategies used in the proposed method may have interfered with this holistic mapping, suggesting that simpler visual reasoning approaches would be more effective.

\textbf{Relationship Between Task Characteristics and Effects}: Overall, the method proved highly effective for tasks requiring step-by-step numerical processing (e.g., Object Counting) and multi-stage reasoning (e.g., Multistep Arithmetic). In contrast, for tasks requiring intuitive visual understanding (e.g., Geometric Shapes), complex strategies sometimes hindered performance. These findings highlight the need to refine technique selection mechanisms to better align with task characteristics.

\subsection{Temperature Optimization Experiments}

\textbf{Experimental Context and Theoretical Background}:  
While prompt content design has been the main focus of LLM performance optimization, runtime parameters also play a critical role in shaping output quality. Among them, the temperature parameter controls the balance between determinism and creativity, with optimal values varying empirically by task type.

\textbf{Research Hypothesis}:  
We hypothesize that ``different task characteristics require different optimal temperature parameters.'' This experiment quantitatively evaluates the impact of task-specific temperature optimization and examines the effectiveness of integrating prompt design with execution parameter tuning.

To test this hypothesis, we conducted temperature optimization experiments to assess whether adjusting parameters for individual tasks yields additional performance gains.

\subsubsection{Optimization Experiment Setup}

\textbf{Temperature Parameter Candidates}:  
Seven values were tested: 0.0, 0.2, 0.4, 0.6, 0.8, 1.0, and 1.3. These cover a wide spectrum of response characteristics, from deterministic to creative outputs.

\textbf{Constraints for Experimental Efficiency}:  
Because optimizing across all problems would be computationally prohibitive, 40 problems were randomly sampled from each task for evaluation.

\textbf{Evaluation Process}:  
For each task–temperature pair, 10 independent trials were conducted, and average scores were computed. The temperature yielding the highest average score was selected as the optimal value for that task.

\subsubsection{Optimization Results and Trend Analysis}

As shown in Table \ref{tab:results}, the proposed method with individually optimized temperature parameters (Ours, temperature-optimized) achieved an arithmetic mean of 28.5 and a harmonic mean of 13.3, further surpassing the default configuration.

\textbf{Temperature Parameter Effect Analysis Based on Statistical Significance}: To rigorously evaluate the effect of temperature parameter optimization, we conducted a one-way ANOVA \citep{fisher1918correlation} across temperature settings for each task, and assessed statistical significance using F-statistics and p-values.

\textbf{Classification by Statistical Significance:}

\begin{itemize}
\item \textbf{Highly Significant (p $<$ 0.001)}: 3 tasks
Zebra Puzzles (F=12.36, p$<$0.001), Web of Lies (F=8.34, p$<$0.001), Hyperbaton (F=4.32, p$<$0.001)
Task group where temperature parameters have extremely strong influence on performance
\item \textbf{Significant (0.001 $\leq$ p $<$ 0.05)}: 3 tasks
Boolean Expressions (F=3.42, p=0.003), Multistep Arithmetic (F=2.51, p=0.023), NYCC (F=3.87, p=0.001)
Moderate influence task group where temperature parameter effects are statistically confirmed
\item \textbf{Non-significant (p $\geq$ 0.05)}: 17 tasks
Temperature parameter influence is not statistically detected in the majority of tasks including Geometric Shapes, Object Counting, etc.
\end{itemize}

\textbf{Task Characteristic Analysis for Statistically Significant Tasks}: Among the six tasks in which temperature effects were statistically significant, clear polarization in the optimal temperature values was observed.

\begin{itemize}
\item \textbf{Low Temperature Type (0.0)}: Zebra Puzzles, Web of Lies, Boolean Expressions, Multistep Arithmetic, NYCC
Task group requiring logical rigor, unique solution identification, and convergent thinking, where consistency assurance through low temperature is effective
\item \textbf{High Temperature Type (1.3)}: Hyperbaton
Tasks requiring interpretation of complex combination rules in multi-dimensional systems of adjective order (opinion, size, age, color, material, nationality, etc.), demanding linguistic intuition difficult to address through mechanical application. Response diversity through high temperature realizes natural word order judgment beyond statistical patterns
\end{itemize}

These results indicate that further performance gains can be achieved through comprehensive optimization encompassing both prompt content and LLM execution parameters. In particular, incorporating mechanisms that adaptively adjust temperature settings based on task characteristics is expected to enable the development of more versatile prompt generation systems.

\section{Conclusion}

This paper presented a novel framework that interprets users' abstract task descriptions, adaptively selects optimal prompting techniques, and automatically generates high-quality prompts from scratch. The core of the method lies in constructing a knowledge base that maps Task Clusters, which classify tasks according to their functional characteristics, to diverse prompting techniques, and in dynamically generating prompts that effectively achieve user objectives based on this foundation.

Experiments using the BBEH dataset demonstrated that the proposed method achieved an arithmetic mean of 28.0 and a harmonic mean of 12.5, outperforming both standard prompts (arithmetic mean of 23.9, harmonic mean of 9.7) and existing automatic prompt generation tools (arithmetic mean of 24.7, harmonic mean of 10.5). Furthermore, with individual temperature optimization, performance further improved to an arithmetic mean of 28.5 and a harmonic mean of 13.3.

Although these results validate the effectiveness of the proposed framework on the BBEH benchmark, several limitations should be acknowledged. The knowledge base constructed in this study is tailored to BBEH task characteristics, and it remains an open question whether this knowledge base can be directly applied to other domains. This point represents an important direction for future investigation.

The primary contribution of this research is the establishment of a scalable framework for the semi-automatic construction of task-to-technique knowledge bases, rather than a single instantiation. A key advantage of this framework is its adaptability and low customization cost across different domains. As demonstrated in this work, both task clustering and the initial mapping of techniques can be largely automated using LLMs. Consequently, expert analysis and design costs for applying this framework to new domains can be significantly reduced, representing a major advantage for practical implementation.

Given its high degree of customizability, the most promising future direction is deployment in domain-specific applications. For example, applying this framework to financial report analysis or manufacturing defect detection could enable the rapid development of high-performance prompt generators specialized for those fields. In this way, the method has the potential to serve as a foundational technology for lowering barriers to LLM utilization across diverse specialized domains and accelerating their adoption.

Looking ahead, several extensions are envisioned. Incorporating mechanisms to dynamically update and optimize knowledge bases based on user feedback and emerging prompting techniques, and integrating functionality that allows LLMs to predict and evaluate prompt effectiveness in advance, would further enhance the framework’s reliability and adaptability. Through such extensions, this research can evolve into a core technology for fostering more seamless and effective interaction between LLMs and human users.

\clearpage

\bibliographystyle{plainnat}
\bibliography{references}

\clearpage

\appendix

\section{Prompt Template Examples}

This section presents examples of prompt templates for the three methods compared in this paper. As described in Section 4. 2. 2, the prompt template placeholders include \verb|{$INPUT}| for the problem statement and \verb|{$FINAL_ANSWER_FORMAT}| for the final answer format specification. In practice, these template placeholders are pre-populated with their respective values before executing answer generation by the LLM.

\subsection{Prompt Template Example - Original}

\begin{lstlisting}
{$INPUT}

{$FINAL_ANSWER_FORMAT}
\end{lstlisting}

\subsection{Prompt Template Example - Anthropic}

\begin{lstlisting}
You will be evaluating arithmetic expressions that use custom operators with special definitions. Here is the expression you need to evaluate:

<expression>
{$INPUT}
</expression>

These expressions use custom arithmetic operators that have specific definitions provided within the expression context. Here are the key rules you need to follow:

Custom Operator Definitions:
- Each custom operator (like ><, @, [], etc.) will have a specific definition that tells you how to compute a op b
- Some operators may be defined in terms of other custom operators
- Always use the exact definitions provided

Composition Rules:
- When you see "a op1 op2 b", this means (a op1 b) op2 b
- For example: "4 +* 5" means (4 + 5) * 5 = 9 * 5 = 45
- For example: "3 -*+ 2" means (3 - 2) *+ 2, which then becomes (1 * 2) + 2 = 4

Evaluation Process:
1. First, identify all custom operators and their definitions
2. Parse the expression following standard order of operations (parentheses first, then left to right for same precedence)
3. When encountering composed operations (op1 op2), apply the composition rule
4. Substitute the definitions of custom operators and evaluate step by step
5. Show all intermediate calculations

Important Notes:
- Pay careful attention to parentheses and operator precedence
- When operators are defined recursively (using other custom operators), substitute definitions step by step
- Double-check each arithmetic calculation
- Be systematic and show every step of your work

Use the scratchpad below to work through the problem step by step, then provide your final numerical answer.

<scratchpad>
[Work through the problem systematically here, showing:
1. Identification of all custom operators and their definitions
2. Parsing of the expression structure
3. Step-by-step evaluation with all intermediate calculations
4. Verification of your final result]
</scratchpad>

{$FINAL_ANSWER_FORMAT}
\end{lstlisting}

\subsection{Prompt Template Example - Ours}

\begin{lstlisting}
<role>
You are a distinguished Professor of Formal Logic and Symbolic Computation. Your expertise lies in interpreting abstract operational systems, parsing complex symbolic expressions, and executing them with flawless, step-by-step precision. You will approach this task with the utmost rigor and clarity, adhering strictly to the provided definitions.
</role>

<task>
You are given a mathematical problem that involves newly defined operators. Your task is to compute the final value of the expression by strictly following the provided rules.

{$INPUT}

To solve this problem, you must follow a structured, two-stage methodology. First, you will analyze the problem and create a detailed plan. Second, you will execute that plan meticulously. The new operator definitions are fundamental axioms for this task; you must use them exactly as stated.

Here is the required process:

### 1. Difficulty Analysis

First, analyze the potential challenges in this specific problem. Consider:

* The complexity of the custom operator definitions.
* The nesting of parentheses and the order of operations.
* The application of the operator composition rule (e.g., `a op1 op2 b`).
* Any potential for misinterpretation or calculation errors.

### 2. Plan

Based on your analysis, create a clear, step-by-step plan to solve the expression. The plan must break down the main expression into a sequence of smaller, manageable calculations.

* Step 1: Explicitly list the definitions of all new operators to confirm your understanding.
* Step 2: Deconstruct the main expression, identifying the evaluation order based on parentheses, starting from the innermost parts.
* Step 3: Outline the sequence of substitutions and calculations required to reach the final answer.

### 3. Execution

Execute your plan step-by-step. Show every calculation and substitution clearly. For each step, state which part of the expression you are evaluating and which operator rule you are applying. Leave no part of the process to implicit reasoning. Show your work in full.

<final_answer_format>
CRITICAL COMPLIANCE REQUIRED
MANDATORY FORMAT - ABSOLUTE ADHERENCE

You MUST provide the final answer in the following format. The automated evaluation system will fail if this format is not followed precisely.

{$FINAL_ANSWER_FORMAT}

</final_answer_format>

This task is a rigorous test of your logical and computational abilities. Precision and accuracy are of the utmost importance. Please consider every detail with extreme care and aim for a perfect, error-free result. Your thoroughness and expertise in symbolic manipulation will lead to a reliable and correct outcome.

\end{lstlisting}

\clearpage

\section{Answer Format Compliance Analysis}

\begin{table}[htbp]
\centering
\caption{Pre-correction Score Table (Including Format Non-compliance)}
\label{tab:precorrection}
\scriptsize
\begin{tabular}{lccccc}
\toprule
task name & Original & Anthropic & Ours & Ours (temperature-optimized) \\
\midrule
BoardgameQA & 41.8 & 42.8 & 43.0 & 44.2 \\
Boolean Expressions & 30.4 & 30.4 & 27.2 & 21.6 \\
Buggy Tables & 3.4 & 4.9 & 3.0 & 4.4 \\
Causal Understanding & 52.8 & 51.4 & 50.6 & 51.8 \\
DisambiguationQA & 49.5 & 45.5 & 47.4 & 41.0 \\
Dyck Languages & 12.1 & 13.8 & 23.9 & 20.8 \\
Geometric Shapes & 33.6 & 25.8 & 17.5 & 18.0 \\
Hyperbaton & 3.8 & 2.2 & 7.0 & 5.9 \\
Linguini & 14.2 & 13.8 & 13.9 & 13.6 \\
Movie Recommendation & 57.4 & 74.0 & 66.0 & 65.2 \\
Multistep Arithmetic & 10.6 & 17.2 & 20.7 & 19.4 \\
NYCC & 13.4 & 12.8 & 12.2 & 11.4 \\
Object Counting & 10.8 & 12.5 & 70.0 & 83.8 \\
Object Properties & 1.0 & 2.8 & 3.0 & 3.4 \\
SARC Triples & 34.4 & 30.6 & 34.4 & 35.3 \\
Shuffled Objects & 11.6 & 4.8 & 2.4 & 2.7 \\
Spatial Reasoning & 16.9 & 19.0 & 30.8 & 27.8 \\
SportQA & 21.0 & 25.0 & 23.0 & 24.4 \\
Temporal Sequences & 0.7 & 0.5 & 0.8 & 0.9 \\
Time Arithmetic & 42.7 & 40.5 & 39.4 & 41.4 \\
Web of Lies & 19.8 & 19.9 & 21.2 & 19.7 \\
Word Sorting & 24.5 & 20.6 & 21.7 & 21.9 \\
Zebra Puzzles & 37.4 & 42.0 & 32.8 & 28.4 \\
\midrule
\textbf{arithmetic mean across tasks} & \textbf{23.6} & \textbf{24.0} & \textbf{26.6} & \textbf{26.4} \\
\textbf{harmonic mean across tasks} & \textbf{9.6} & \textbf{9.9} & \textbf{10.9} & \textbf{11.3} \\
\bottomrule
\end{tabular}
\end{table}

\vspace{0.5cm}

\begin{table}[htbp]
\centering
\caption{Detailed Table of Format Non-compliance Rates by Method and Task}
\label{tab:noncompliance}
\scriptsize
\begin{tabular}{lccccc}
\toprule
task name & Original & Anthropic & Ours & Ours (temperature-optimized) \\
\midrule
BoardgameQA & 1.0 & 0.0 & 1.6 & 4.2 \\
Boolean Expressions & 0.4 & 0.7 & 6.6 & 32.5 \\
Buggy Tables & 1.8 & 10.4 & 16.6 & 10.6 \\
Causal Understanding & 0.4 & 0.2 & 0.8 & 0.2 \\
DisambiguationQA & 0.2 & 0.0 & 0.0 & 0.2 \\
Dyck Languages & 1.0 & 0.2 & 0.2 & 0.0 \\
Geometric Shapes & 0.2 & 0.4 & 0.5 & 0.6 \\
Hyperbaton & 0.1 & 0.0 & 0.0 & 0.0 \\
Linguini & 0.7 & 4.6 & 0.2 & 2.3 \\
Movie Recommendation & 0.0 & 0.0 & 7.4 & 0.0 \\
Multistep Arithmetic & 1.0 & 1.0 & 6.6 & 6.4 \\
NYCC & 0.0 & 0.0 & 0.0 & 0.0 \\
Object Counting & 0.0 & 0.0 & 0.0 & 0.2 \\
Object Properties & 0.8 & 0.9 & 18.8 & 24.7 \\
SARC Triples & 1.8 & 0.3 & 0.0 & 0.0 \\
Shuffled Objects & 27.5 & 68.2 & 68.2 & 75.1 \\
Spatial Reasoning & 0.7 & 0.0 & 13.3 & 13.8 \\
SportQA & 1.8 & 0.0 & 0.0 & 0.4 \\
Temporal Sequences & 0.2 & 0.1 & 2.2 & 8.0 \\
Time Arithmetic & 0.2 & 1.1 & 0.0 & 0.4 \\
Web of Lies & 0.4 & 0.5 & 0.4 & 6.3 \\
Word Sorting & 0.4 & 0.0 & 0.2 & 0.9 \\
Zebra Puzzles & 1.0 & 1.6 & 19.0 & 36.2 \\
\bottomrule
\end{tabular}
\end{table}

\clearpage

\section{Box Plots of Score Distributions}

Box plots showing the distribution of accuracy rates obtained from 10 independent trials for each task and each method, presented by task.

\begin{figure}[htbp]
\centering
\includegraphics[width=0.9\textwidth]{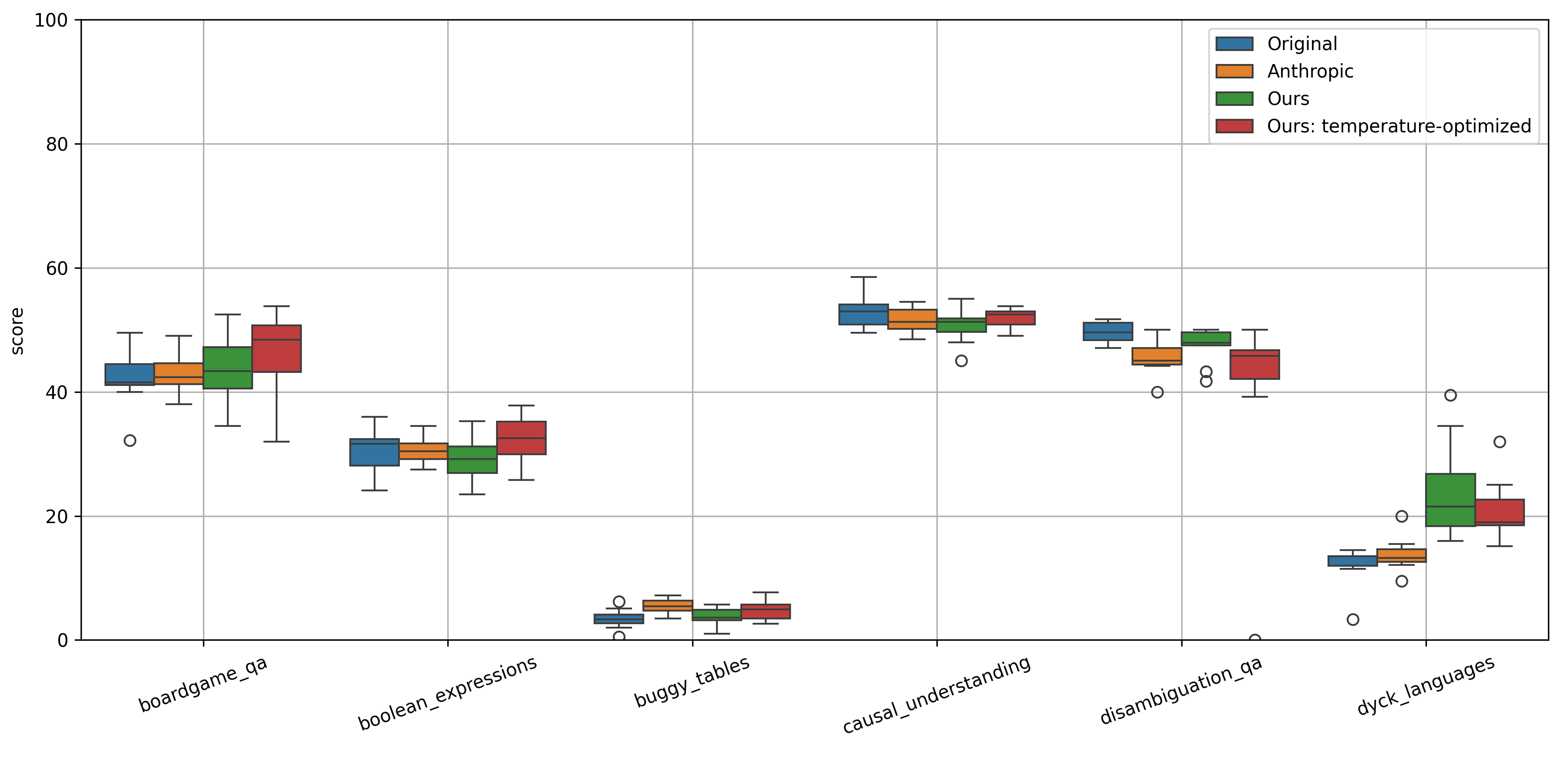}
\caption{Box plots of score distributions (Tasks 1-6)}
\label{fig:boxplot_0}
\end{figure}

\begin{figure}[htbp]
\centering
\includegraphics[width=0.9\textwidth]{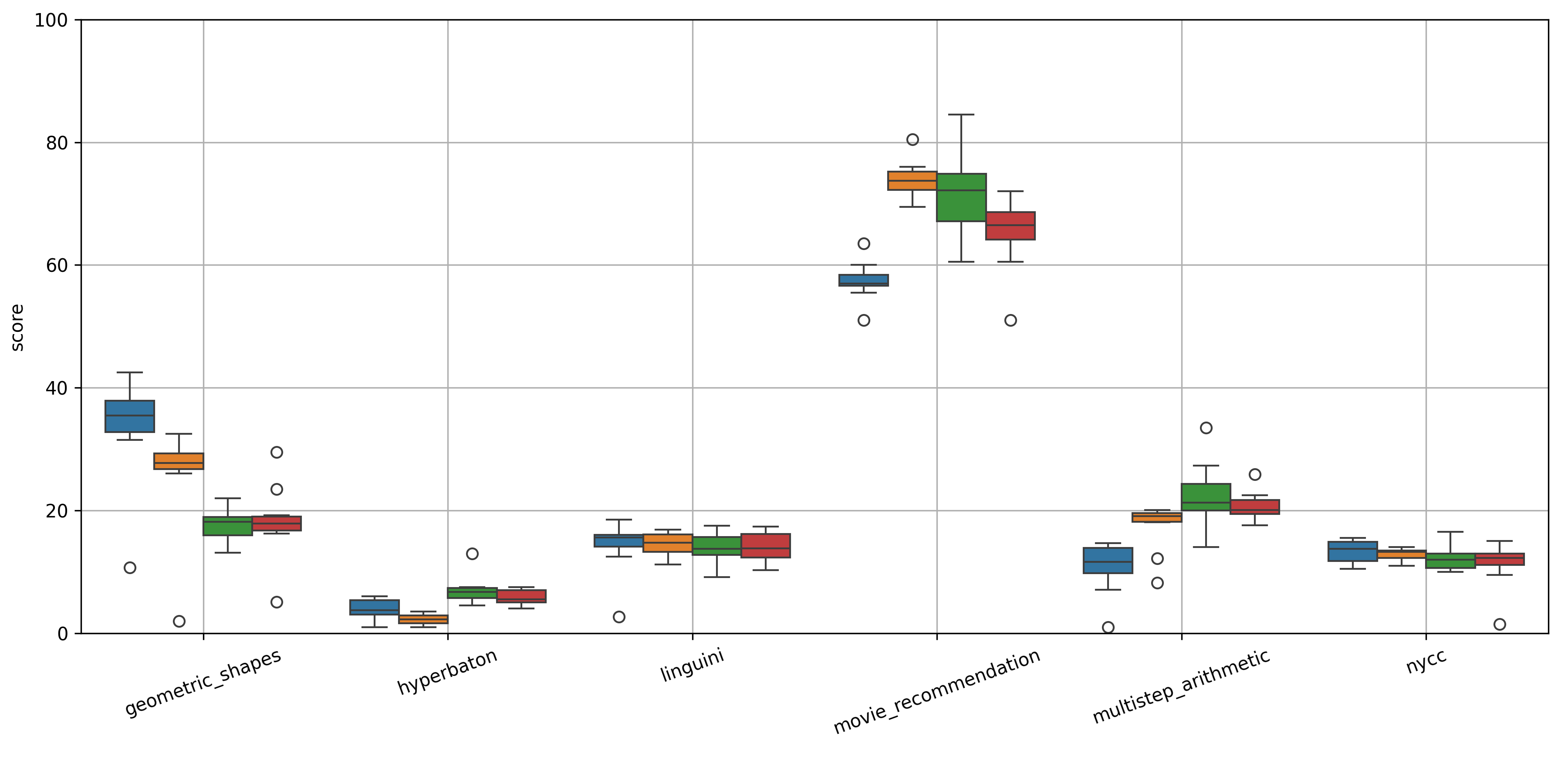}
\caption{Box plots of score distributions (Tasks 7-12)}
\label{fig:boxplot_1}
\end{figure}

\begin{figure}[htbp]
\centering
\includegraphics[width=0.9\textwidth]{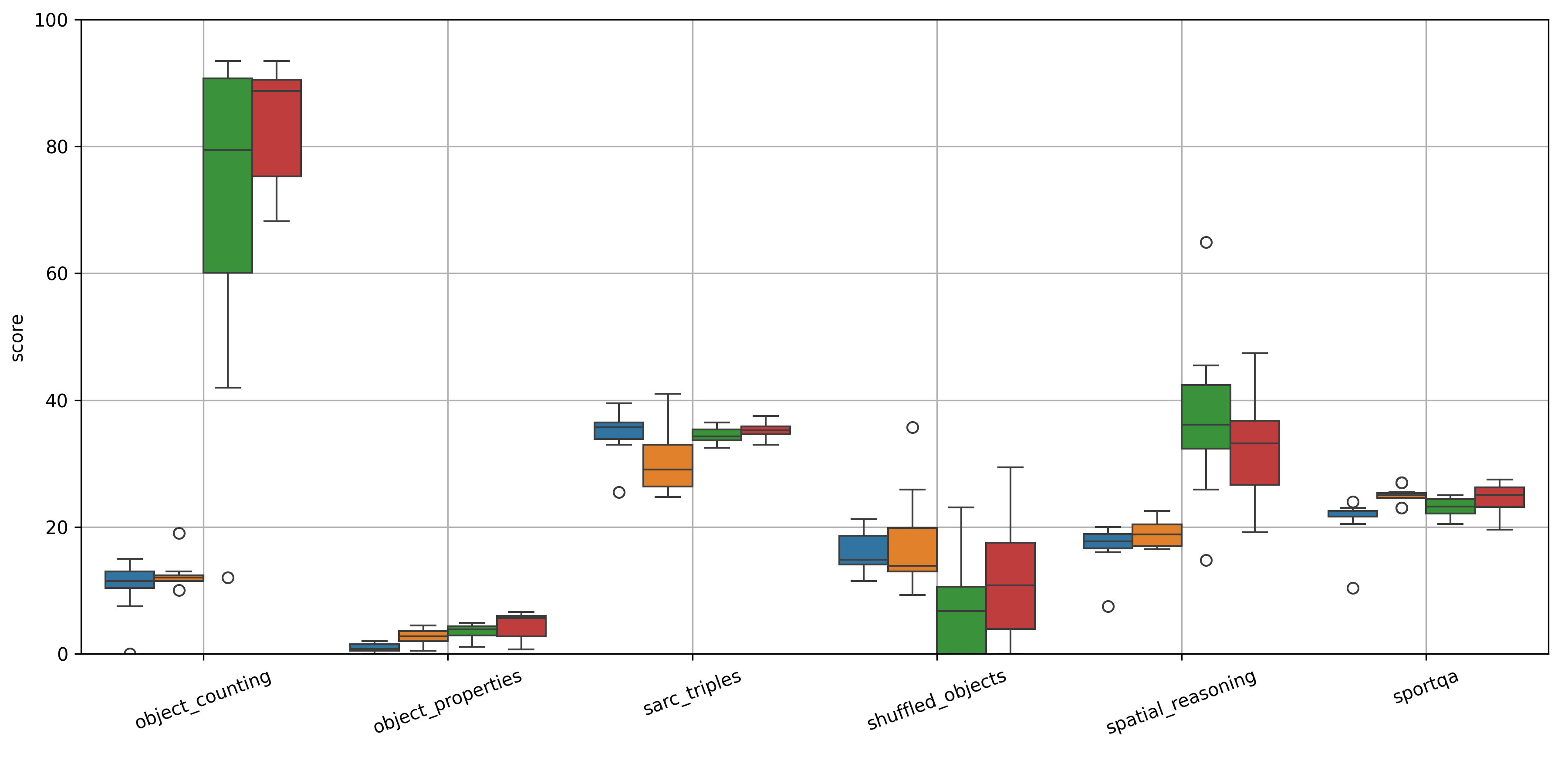}
\caption{Box plots of score distributions (Tasks 13-18)}
\label{fig:boxplot_2}
\end{figure}

\begin{figure}[htbp]
\centering
\includegraphics[width=0.9\textwidth]{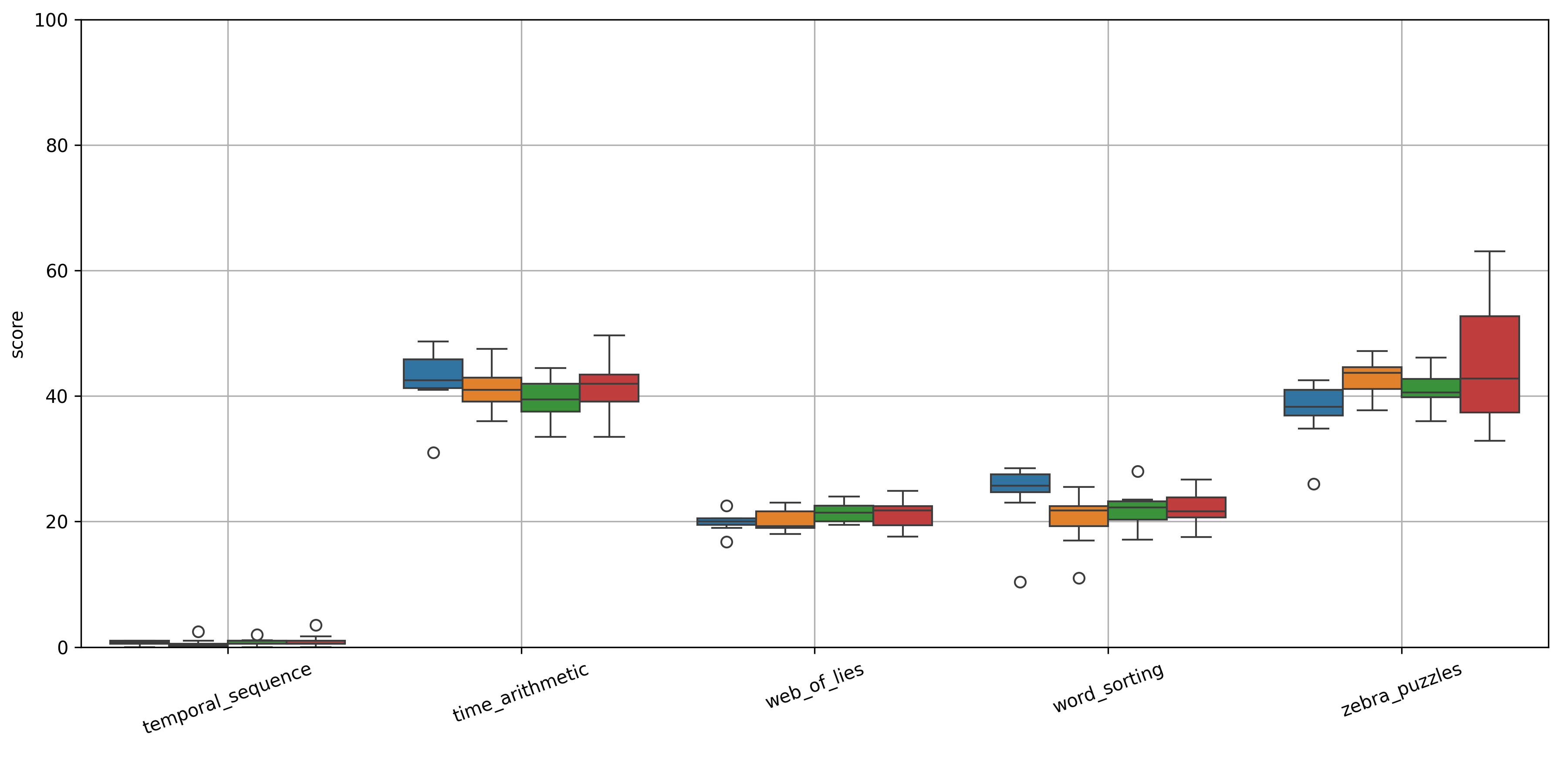}
\caption{Box plots of score distributions (Tasks 19-23)}
\label{fig:boxplot_3}
\end{figure}

\FloatBarrier
\clearpage

\section{Temperature Parameter Optimization Results}

\begin{table}[htbp]
\centering
\caption{Optimal Temperature Parameters for Each Task and ANOVA-based Significance Analysis}
\label{tab:temperature_anova}
\scriptsize
\begin{longtable}{lcccc}
\toprule
task name & optimal temperature & optimal mean score & F statistic value & ANOVA p value \\
\midrule
BoardgameQA & 0.2 & 45.87 & 1.93 & 0.078 \\
Boolean Expressions & 0.0 & 33.73 & 3.42 & 0.003 \\
Buggy Tables & 0.6 & 3.24 & 0.85 & 0.531 \\
Causal Understanding & 1.0 & 49.78 & 1.40 & 0.215 \\
DisambiguationQA & 0.4 & 41.29 & 0.97 & 0.446 \\
Dyck Languages & 1.3 & 19.06 & 1.47 & 0.189 \\
Geometric Shapes & 0.0 & 21.29 & 0.28 & 0.948 \\
Hyperbaton & 1.3 & 8.84 & 4.32 & $<$ 0.001 \\
Linguini & 0.2 & 24.10 & 0.85 & 0.53 \\
Movie Recommendation & 0.8 & 74.94 & 0.64 & 0.698 \\
Multistep Arithmetic & 0.0 & 28.20 & 2.51 & 0.023 \\
NYCC & 0.0 & 15.00 & 3.87 & 0.001 \\
Object Counting & 0.2 & 88.04 & 0.32 & 0.926 \\
Object Properties & 0.2 & 5.27 & 1.64 & 0.139 \\
SARC Triples & 0.2 & 49.42 & 1.07 & 0.385 \\
Shuffled Objects & 1.0 & 9.78 & 1.61 & 0.147 \\
Spatial Reasoning & 1.3 & 32.00 & 1.32 & 0.248 \\
SportQA & 0.2 & 37.53 & 0.37 & 0.896 \\
Temporal Sequences & 0.4 & 1.54 & 1.10 & 0.361 \\
Time Arithmetic & 1.0 & 45.83 & 0.65 & 0.692 \\
Web of Lies & 0.0 & 19.82 & 8.34 & $<$ 0.001 \\
Word Sorting & 1.0 & 17.12 & 1.64 & 0.137 \\
Zebra Puzzles & 0.0 & 58.83 & 12.36 & $<$ 0.001 \\
\bottomrule
\end{longtable}
\end{table}

\FloatBarrier

\clearpage

\section{Temperature Parameter Detailed Visualizations}

The following presents score distributions and average scores for each temperature parameter across tasks.
Note that trial01-03 in the figures below represent results calculated independently using different prompt generation outcomes.

\begin{itemize}
\item Left: Score distribution for each temperature
\item Right: Average scores by temperature with 95\% confidence intervals
\end{itemize}

\begin{figure}[htbp]
\centering
\includegraphics[width=0.9\textwidth]{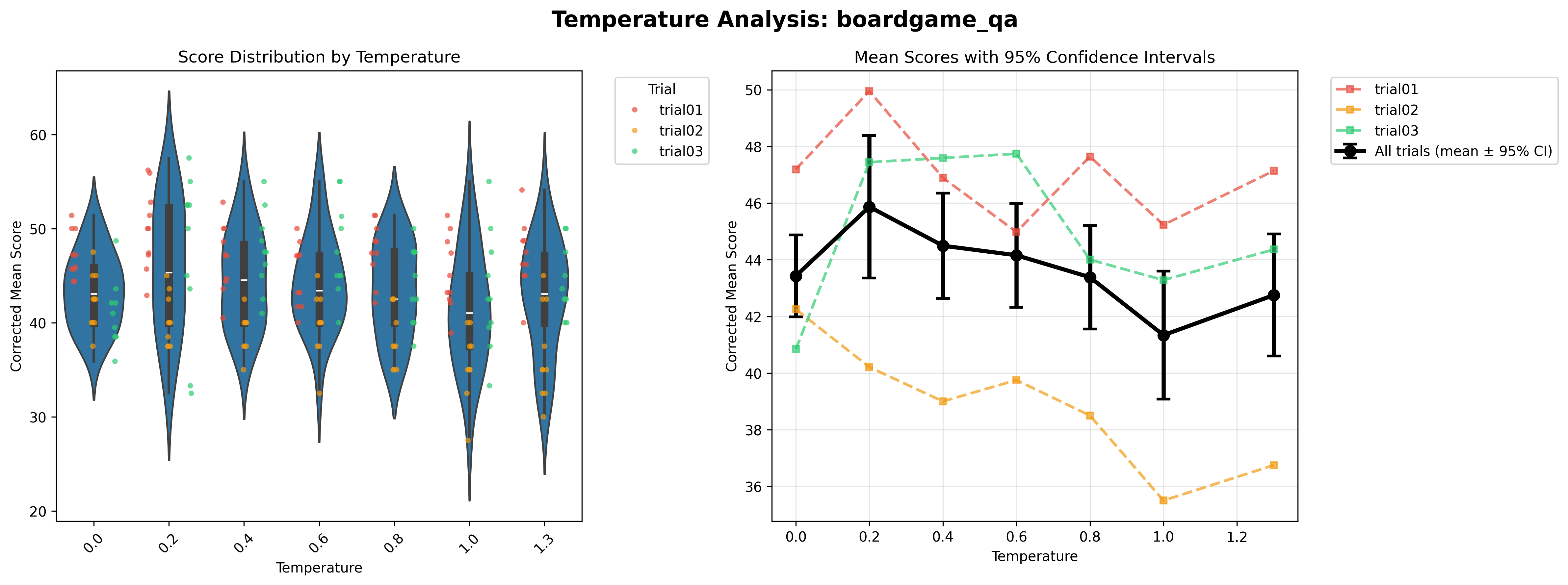}
\caption{Temperature parameter analysis for BoardgameQA}
\label{fig:temp_boardgame}
\end{figure}

\begin{figure}[htbp]
\centering
\includegraphics[width=0.9\textwidth]{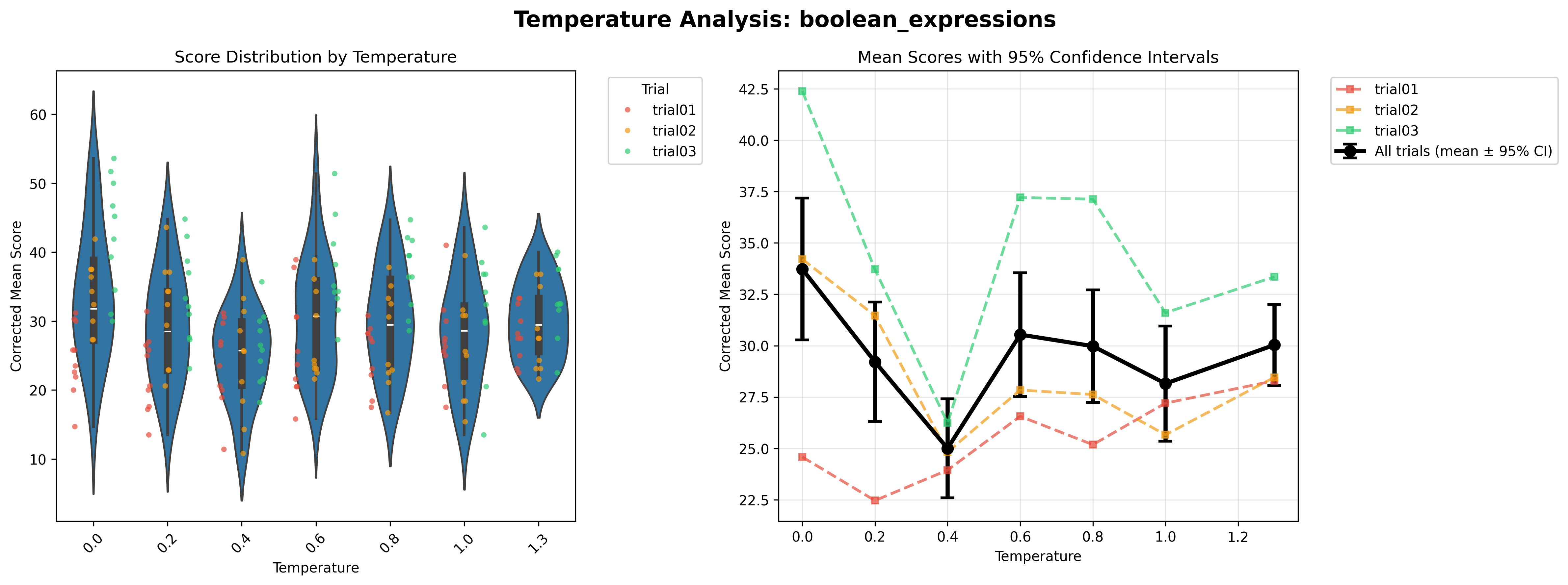}
\caption{Temperature parameter analysis for Boolean Expressions}
\label{fig:temp_boolean}
\end{figure}

\begin{figure}[htbp]
\centering
\includegraphics[width=0.9\textwidth]{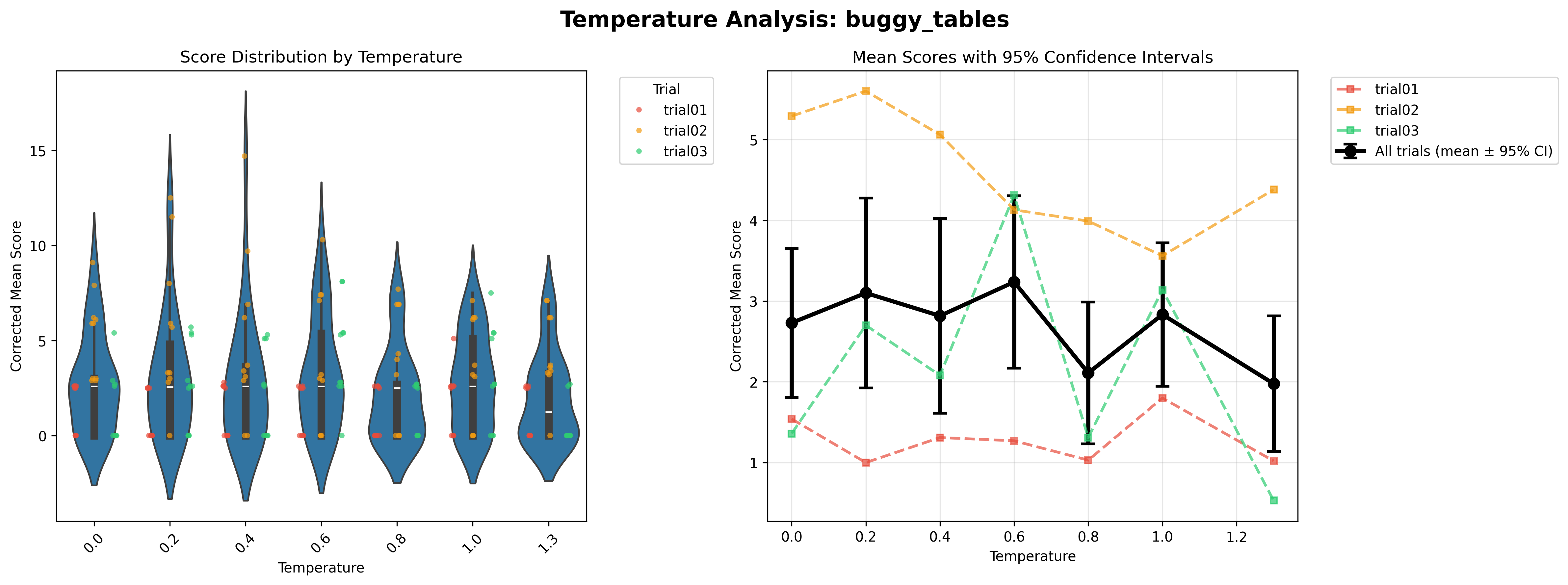}
\caption{Temperature parameter analysis for Buggy Tables}
\label{fig:temp_buggy}
\end{figure}

\begin{figure}[htbp]
\centering
\includegraphics[width=0.9\textwidth]{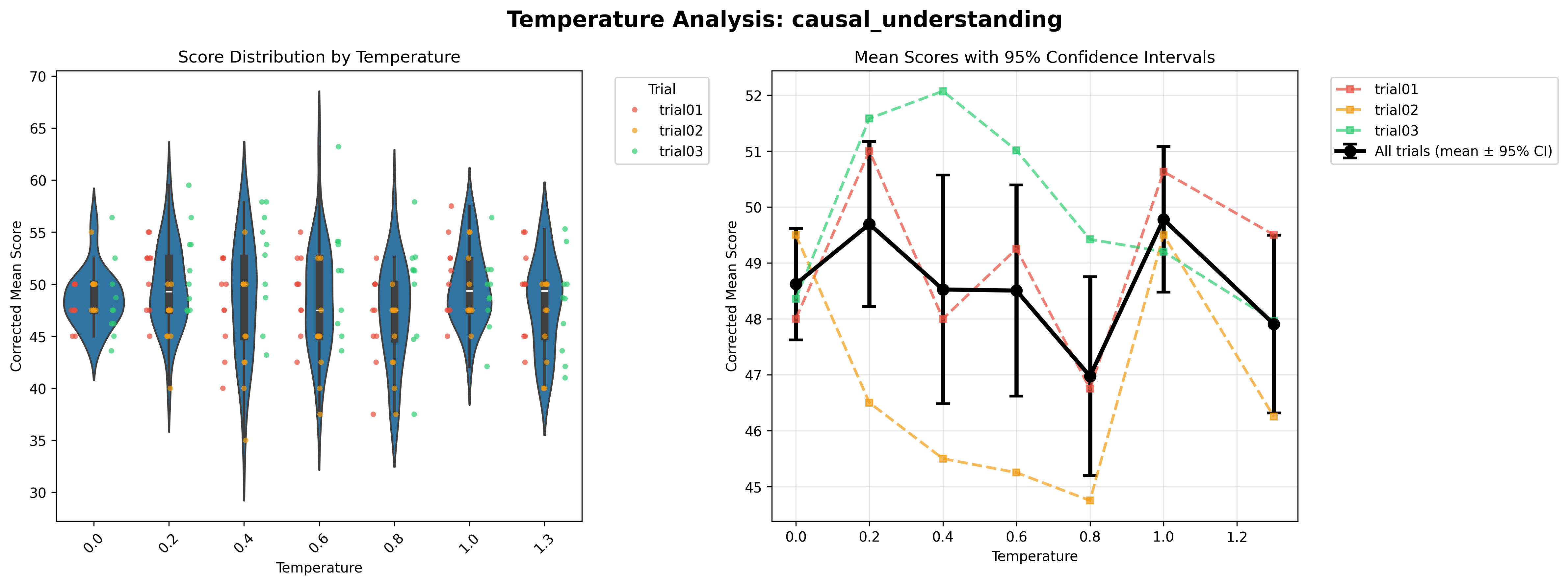}
\caption{Temperature parameter analysis for Causal Understanding}
\label{fig:temp_causal}
\end{figure}

\begin{figure}[htbp]
\centering
\includegraphics[width=0.9\textwidth]{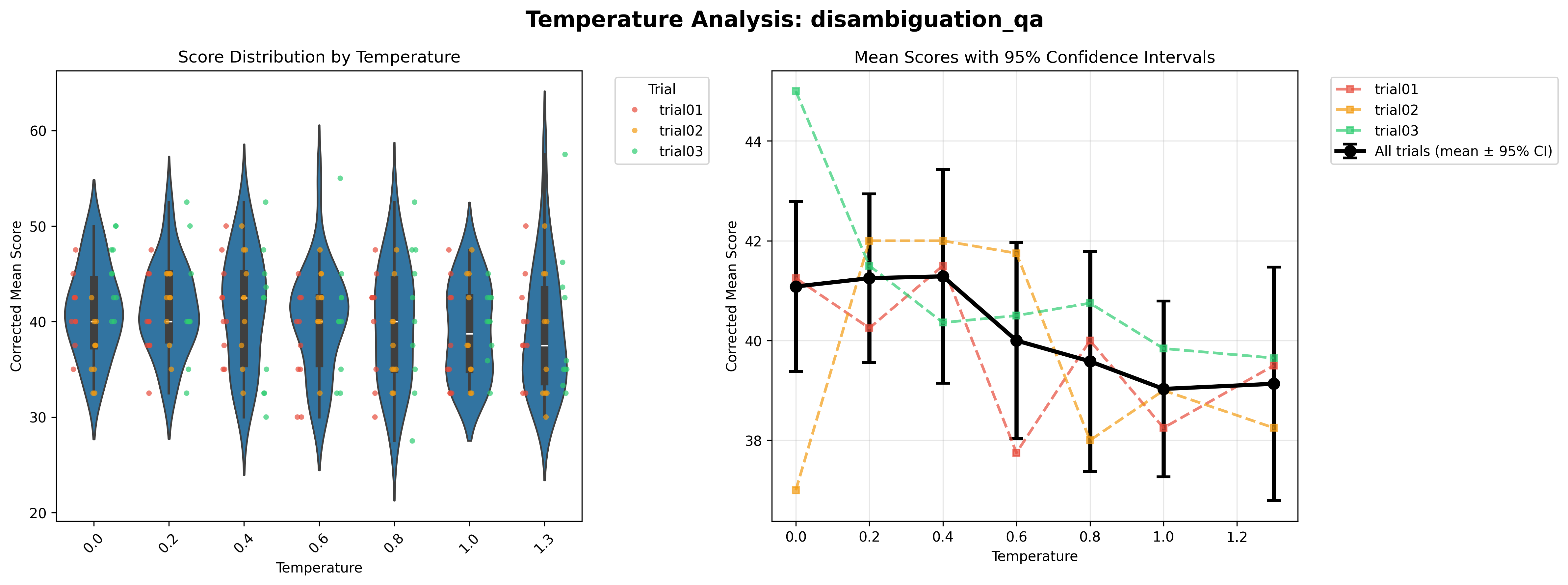}
\caption{Temperature parameter analysis for DisambiguationQA}
\label{fig:temp_disambiguation}
\end{figure}

\begin{figure}[htbp]
\centering
\includegraphics[width=0.9\textwidth]{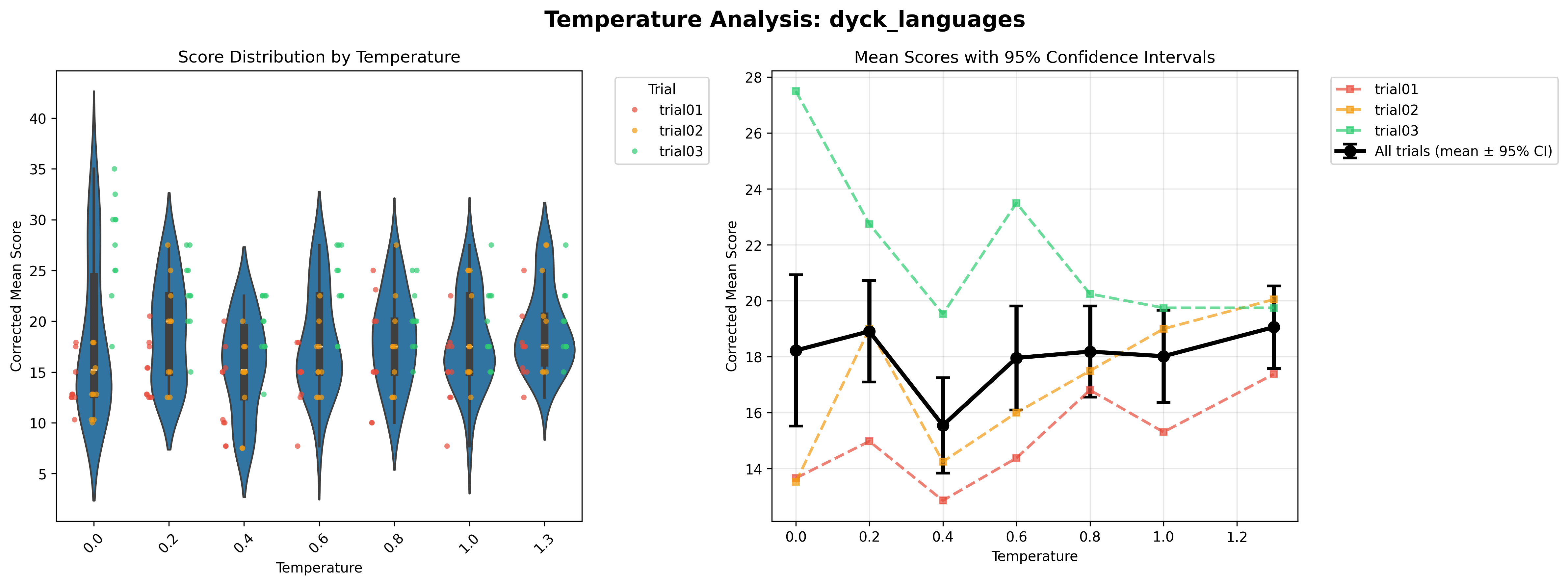}
\caption{Temperature parameter analysis for Dyck Languages}
\label{fig:temp_dyck}
\end{figure}

\begin{figure}[htbp]
\centering
\includegraphics[width=0.9\textwidth]{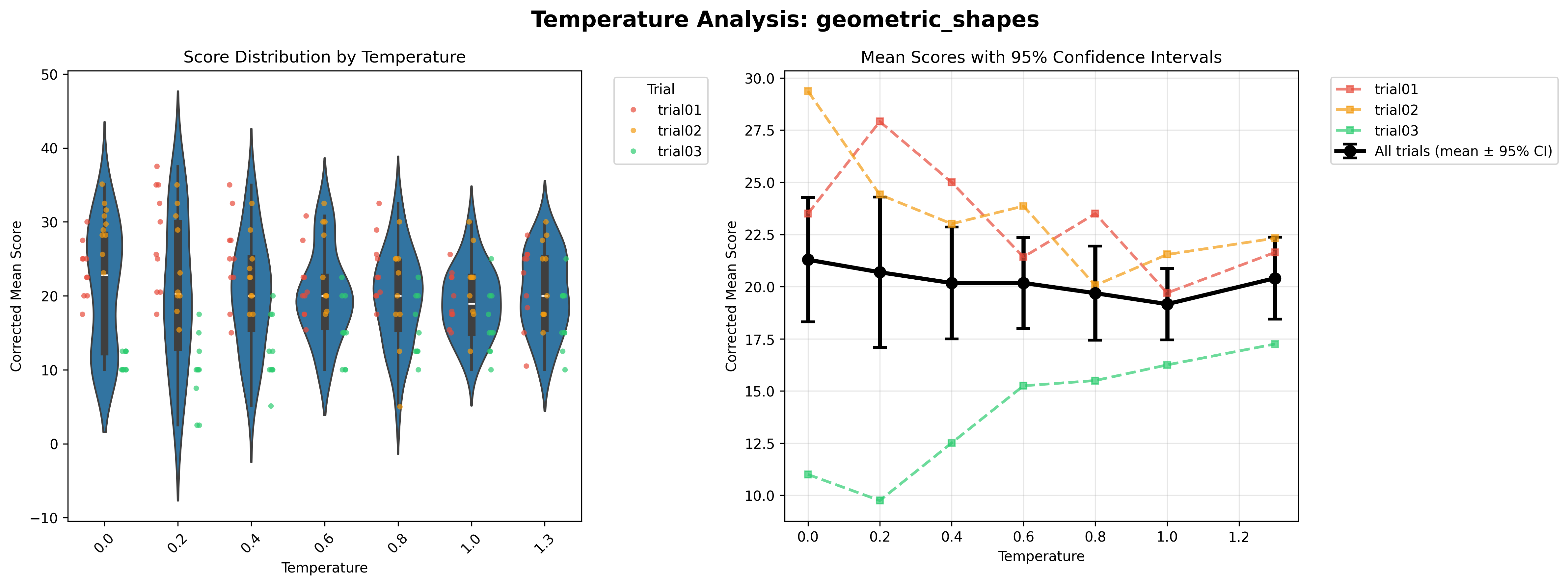}
\caption{Temperature parameter analysis for Geometric Shapes}
\label{fig:temp_geometric}
\end{figure}

\begin{figure}[htbp]
\centering
\includegraphics[width=0.9\textwidth]{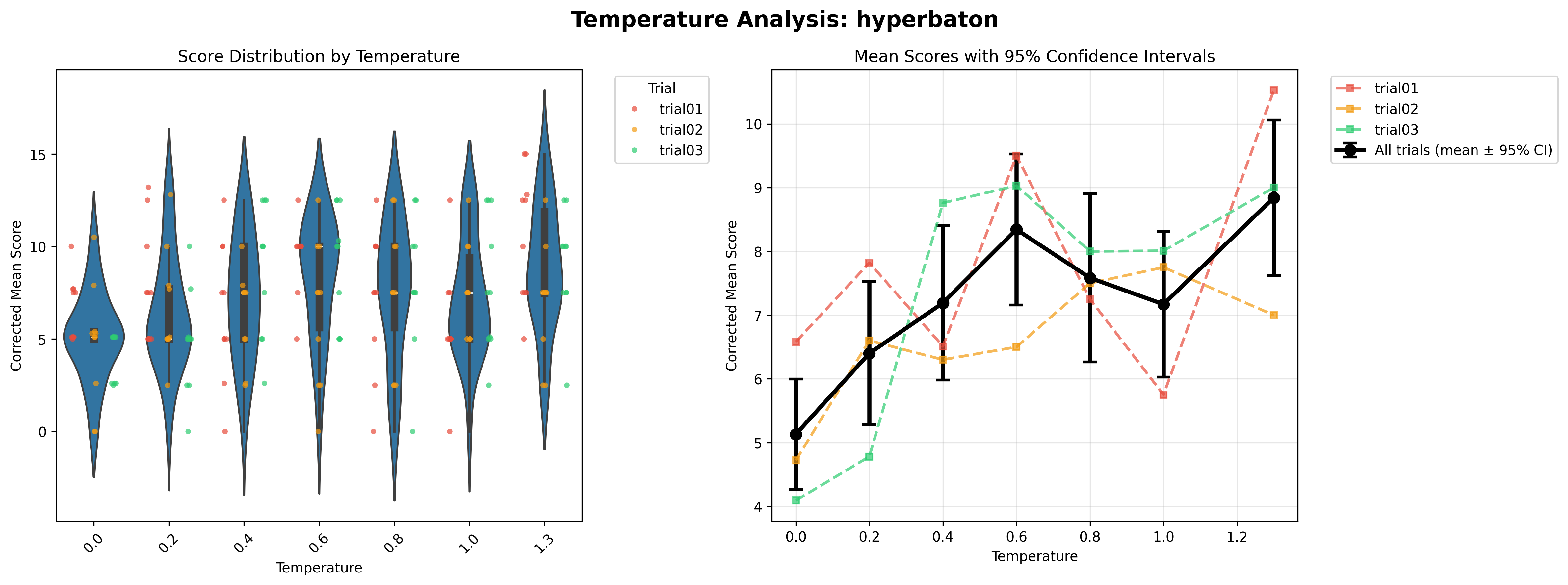}
\caption{Temperature parameter analysis for Hyperbaton}
\label{fig:temp_hyperbaton}
\end{figure}

\begin{figure}[htbp]
\centering
\includegraphics[width=0.9\textwidth]{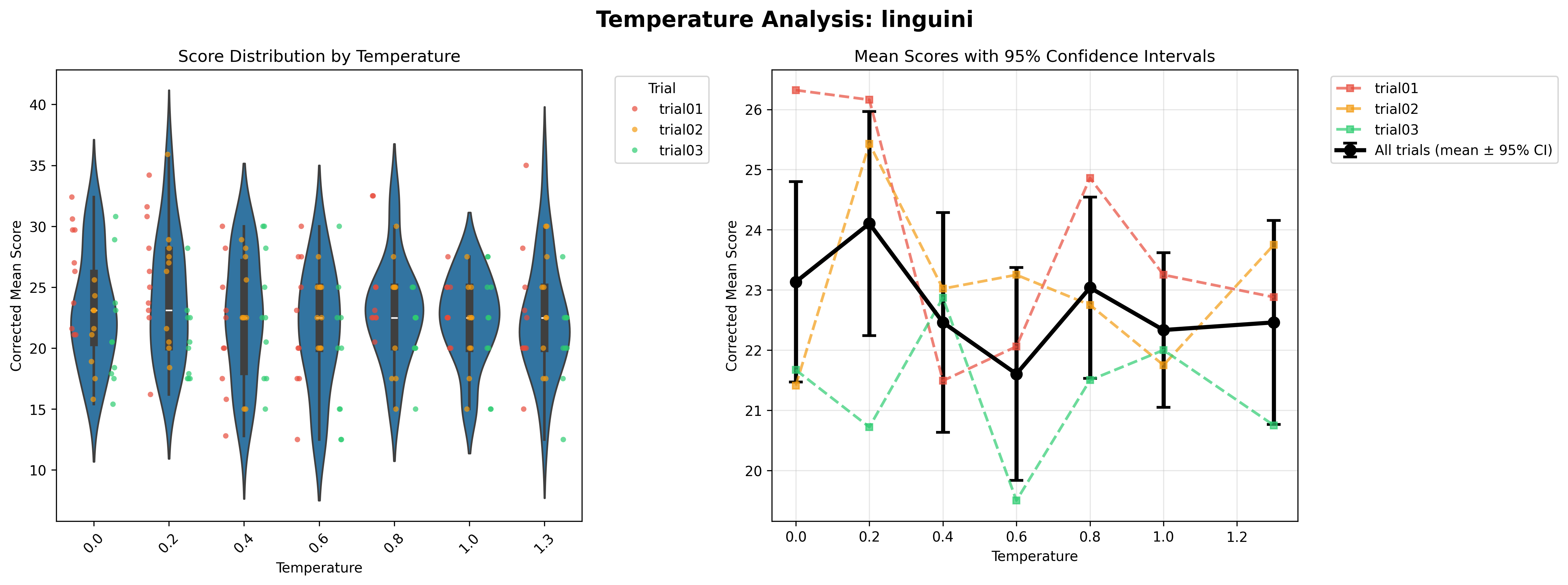}
\caption{Temperature parameter analysis for Linguini}
\label{fig:temp_linguini}
\end{figure}

\begin{figure}[htbp]
\centering
\includegraphics[width=0.9\textwidth]{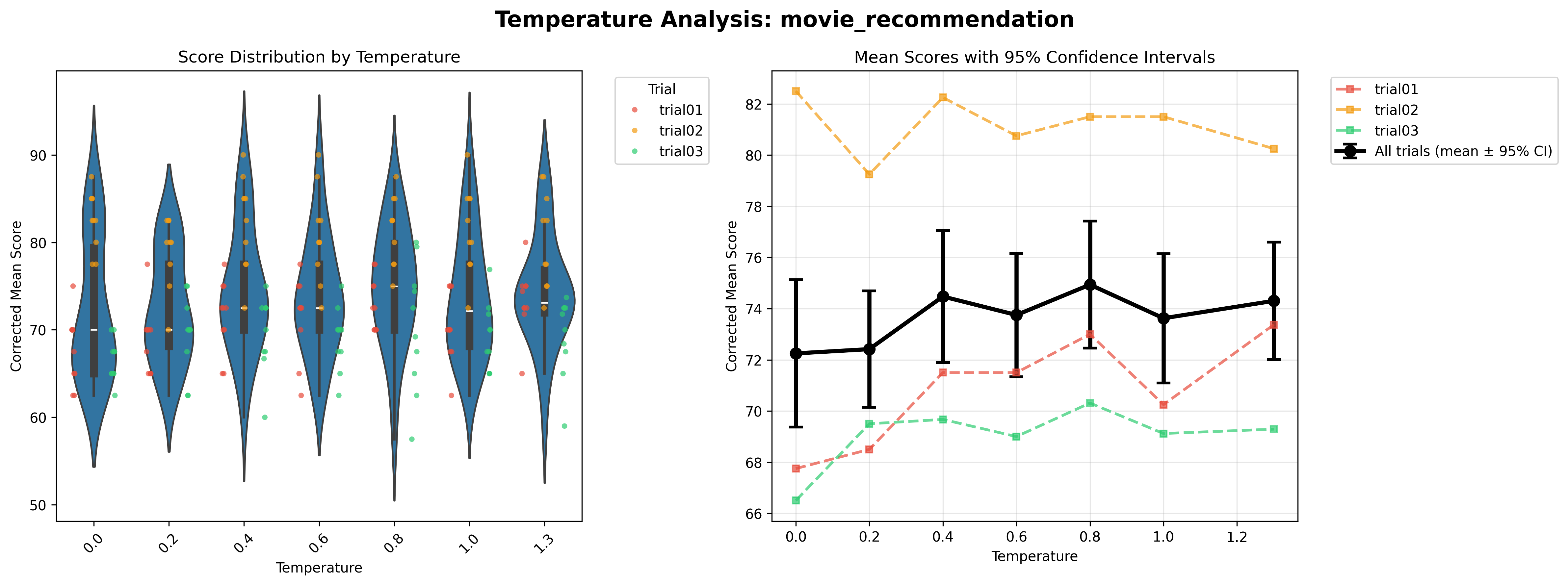}
\caption{Temperature parameter analysis for Movie Recommendation}
\label{fig:temp_movie}
\end{figure}

\begin{figure}[htbp]
\centering
\includegraphics[width=0.9\textwidth]{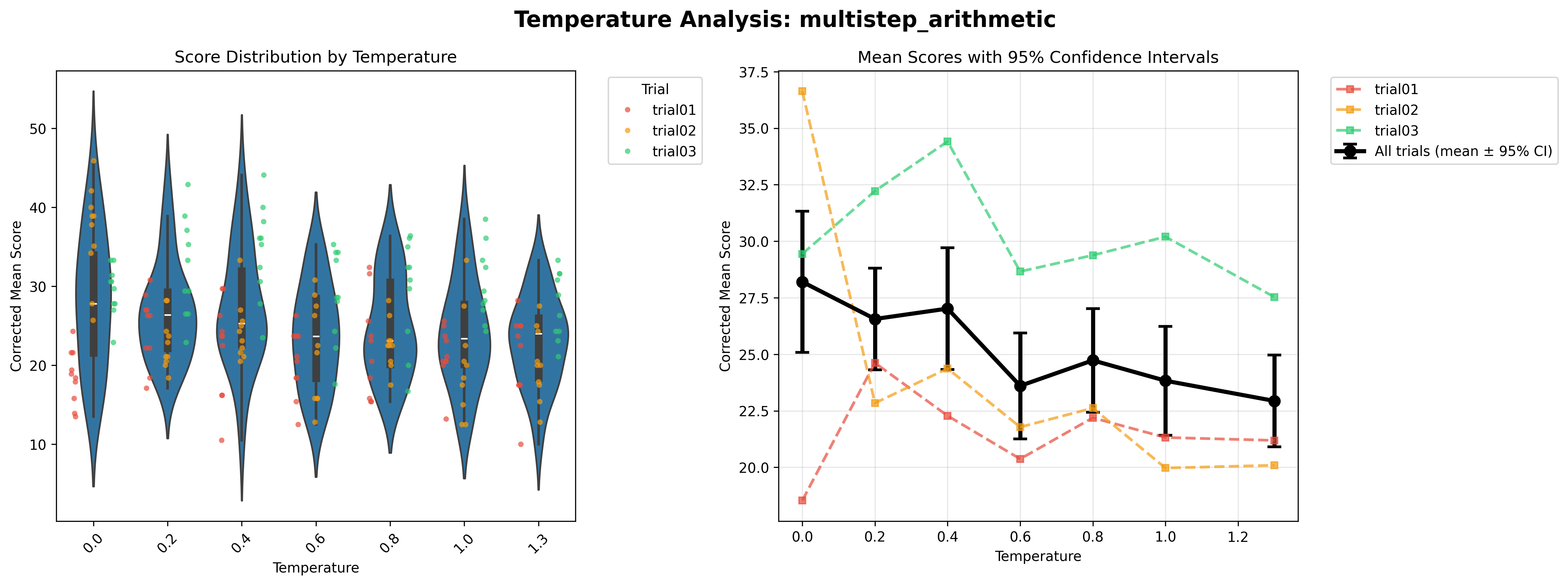}
\caption{Temperature parameter analysis for Multistep Arithmetic}
\label{fig:temp_multistep}
\end{figure}

\begin{figure}[htbp]
\centering
\includegraphics[width=0.9\textwidth]{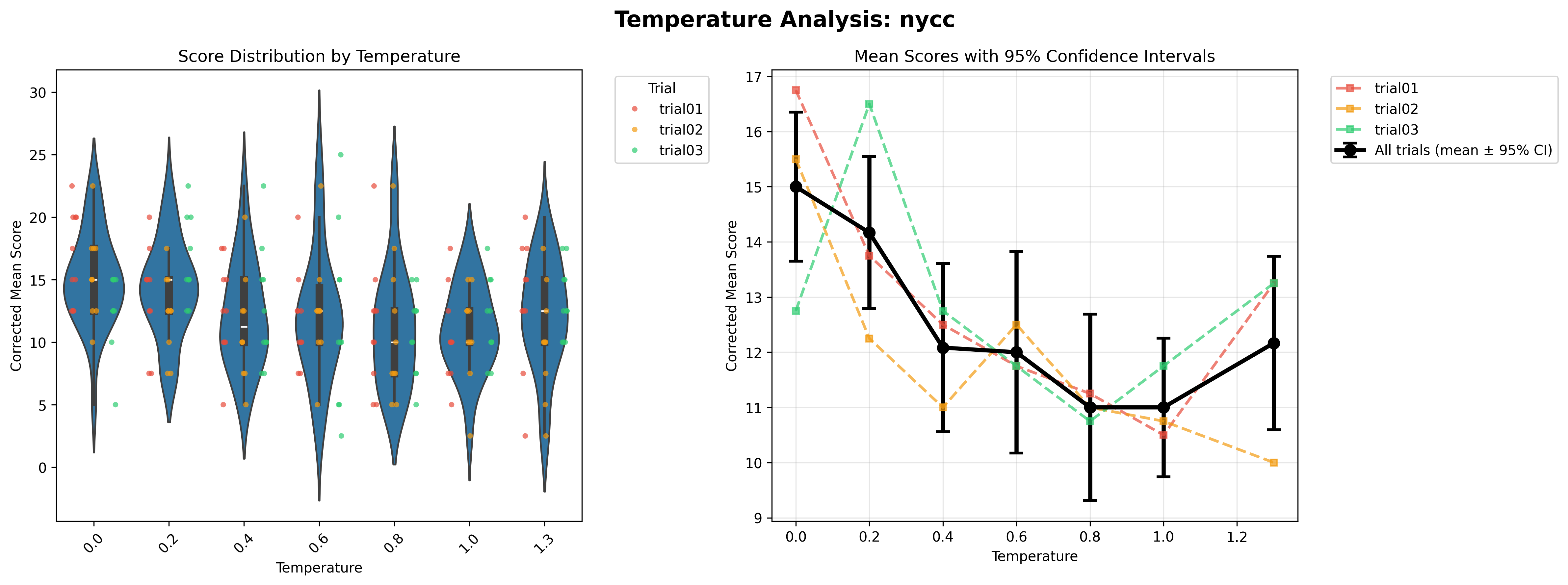}
\caption{Temperature parameter analysis for NYCC}
\label{fig:temp_nycc}
\end{figure}

\begin{figure}[htbp]
\centering
\includegraphics[width=0.9\textwidth]{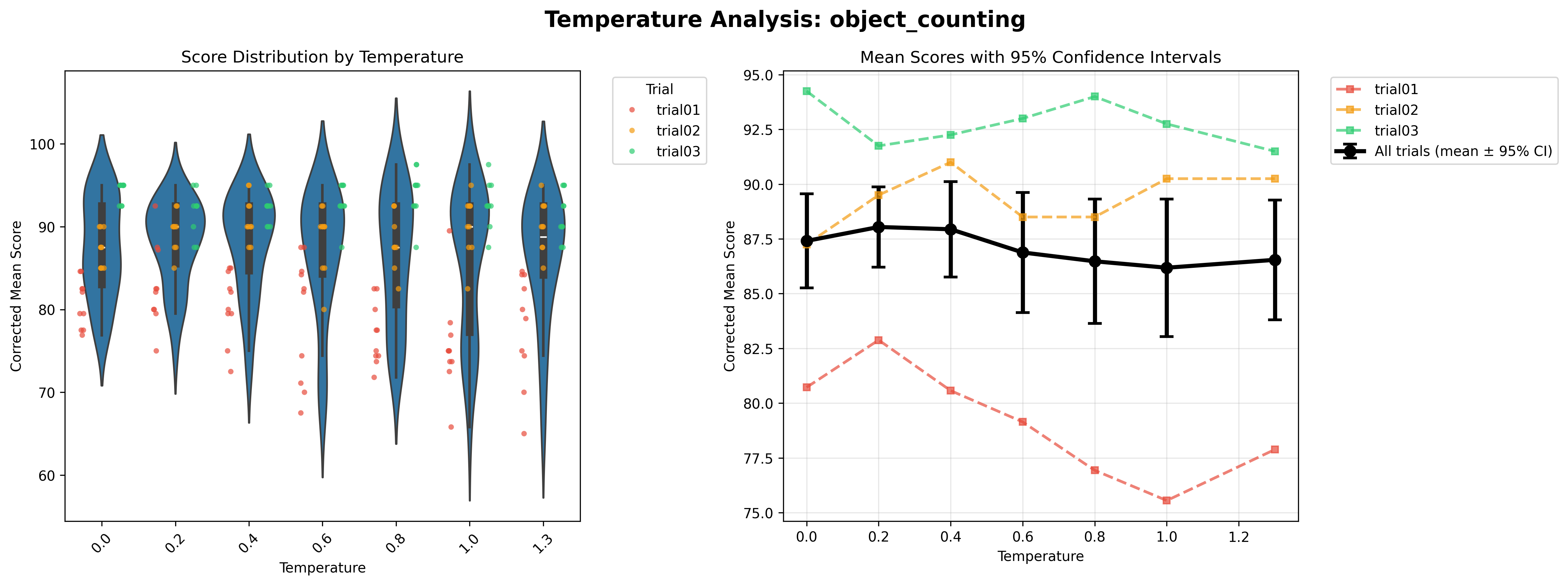}
\caption{Temperature parameter analysis for Object Counting}
\label{fig:temp_object_counting}
\end{figure}

\begin{figure}[htbp]
\centering
\includegraphics[width=0.9\textwidth]{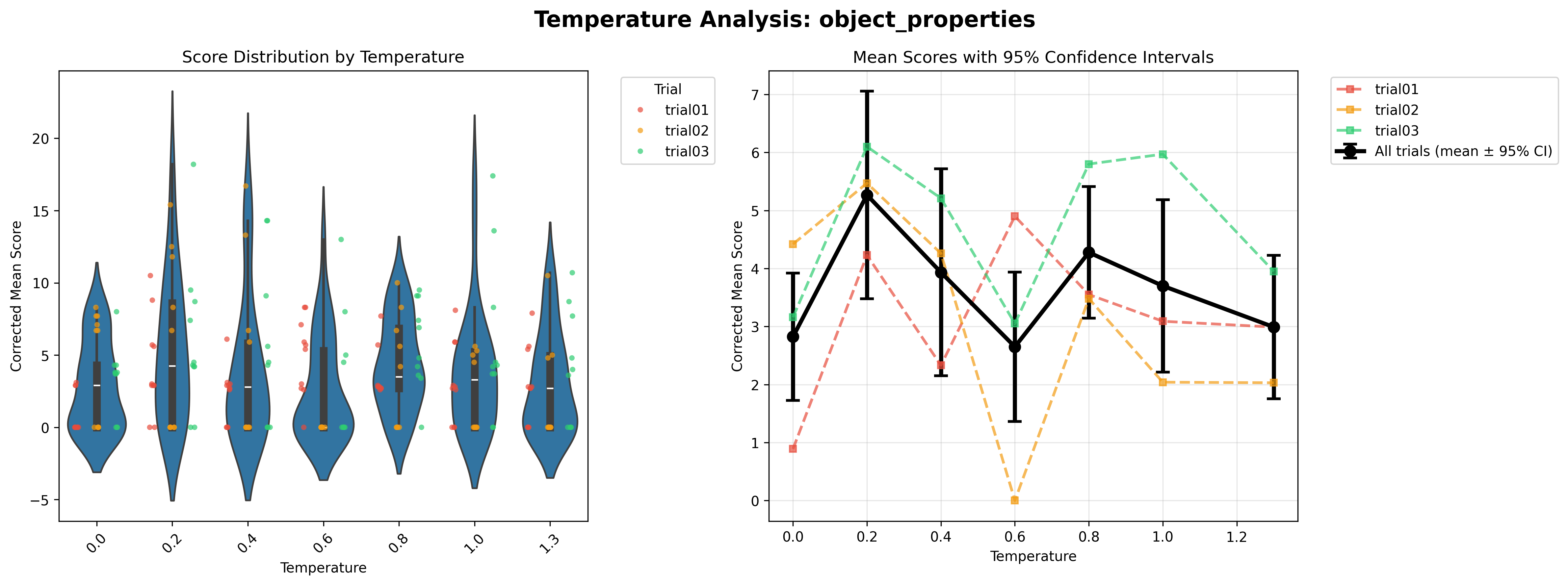}
\caption{Temperature parameter analysis for Object Properties}
\label{fig:temp_object_properties}
\end{figure}

\begin{figure}[htbp]
\centering
\includegraphics[width=0.9\textwidth]{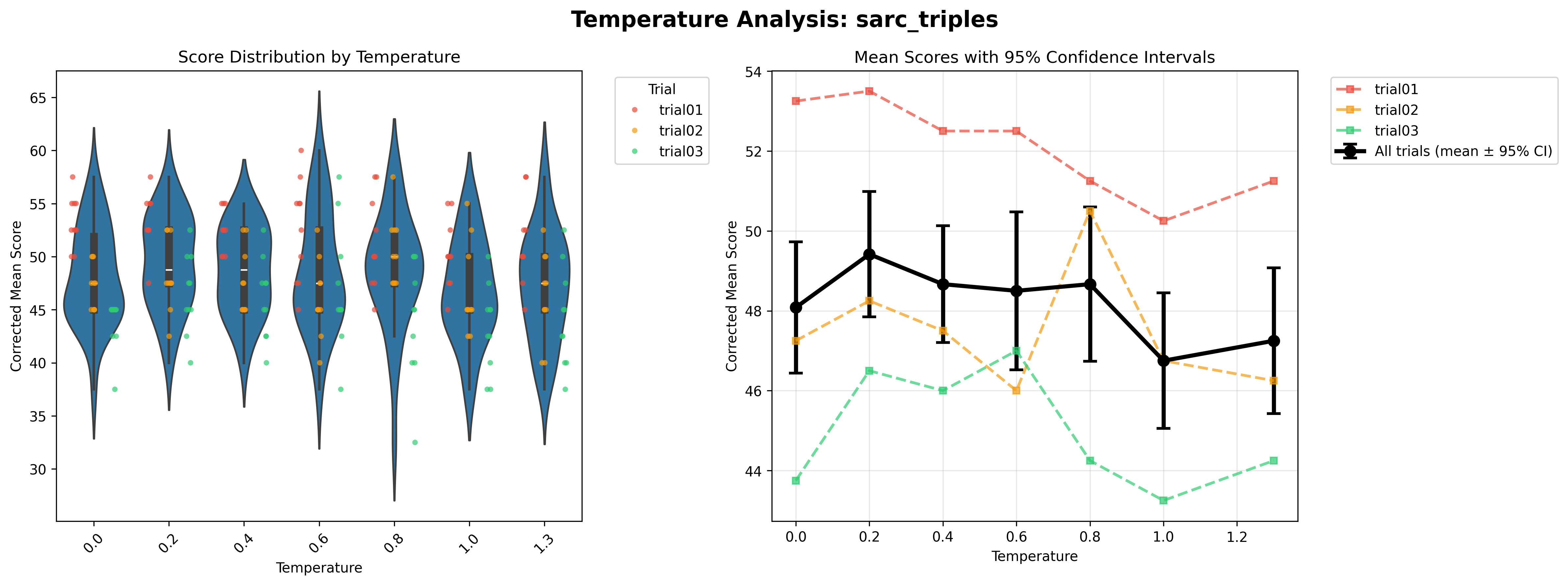}
\caption{Temperature parameter analysis for SARC Triples}
\label{fig:temp_sarc}
\end{figure}

\begin{figure}[htbp]
\centering
\includegraphics[width=0.9\textwidth]{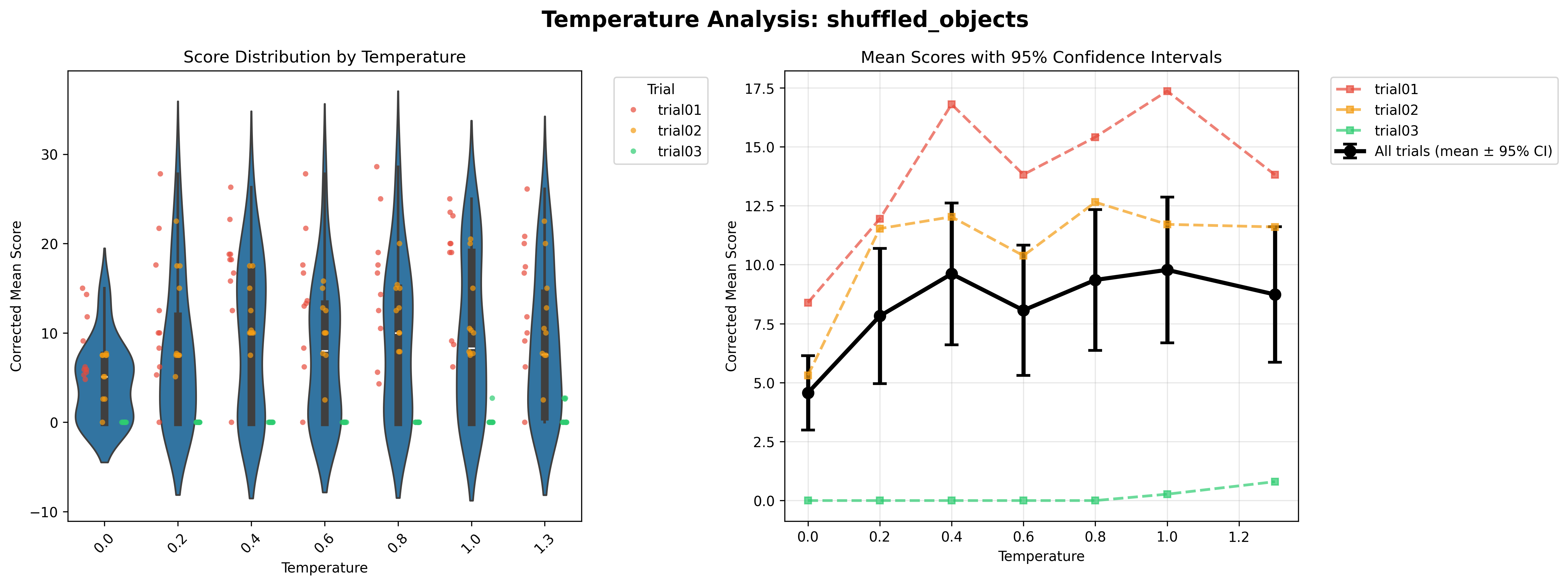}
\caption{Temperature parameter analysis for Shuffled Objects}
\label{fig:temp_shuffled}
\end{figure}

\begin{figure}[htbp]
\centering
\includegraphics[width=0.9\textwidth]{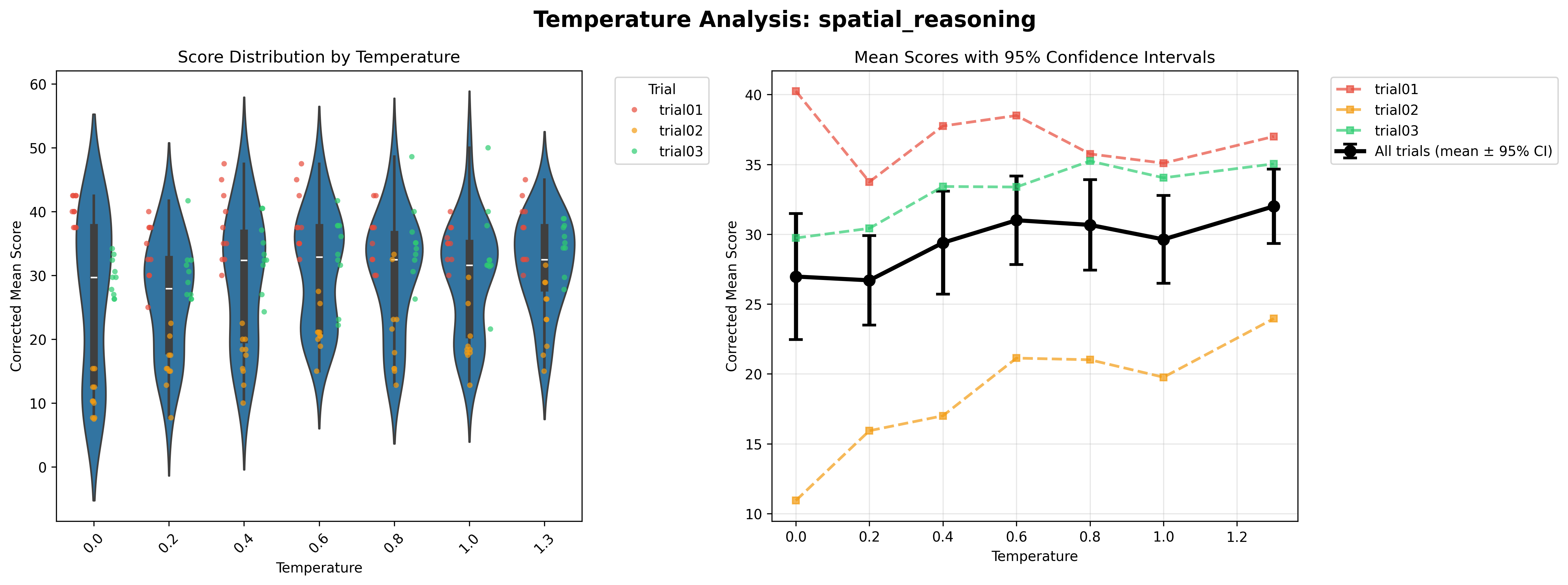}
\caption{Temperature parameter analysis for Spatial Reasoning}
\label{fig:temp_spatial}
\end{figure}

\begin{figure}[htbp]
\centering
\includegraphics[width=0.9\textwidth]{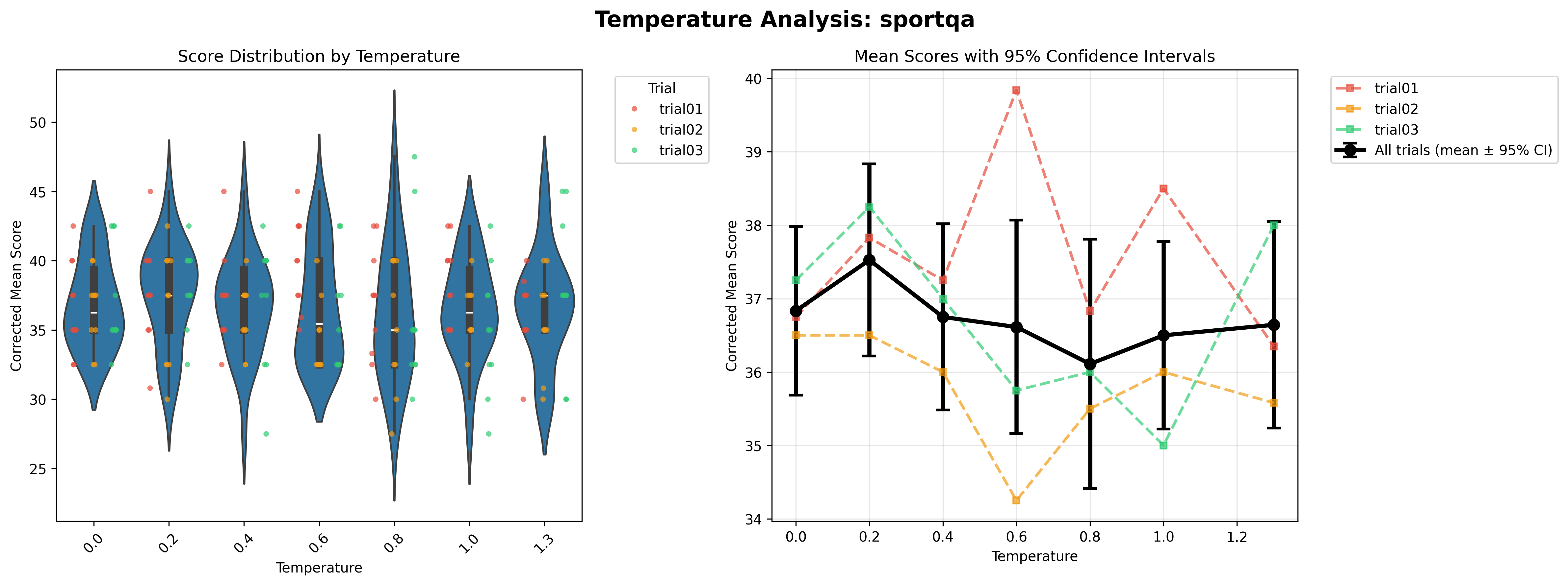}
\caption{Temperature parameter analysis for SportQA}
\label{fig:temp_sport}
\end{figure}

\begin{figure}[htbp]
\centering
\includegraphics[width=0.9\textwidth]{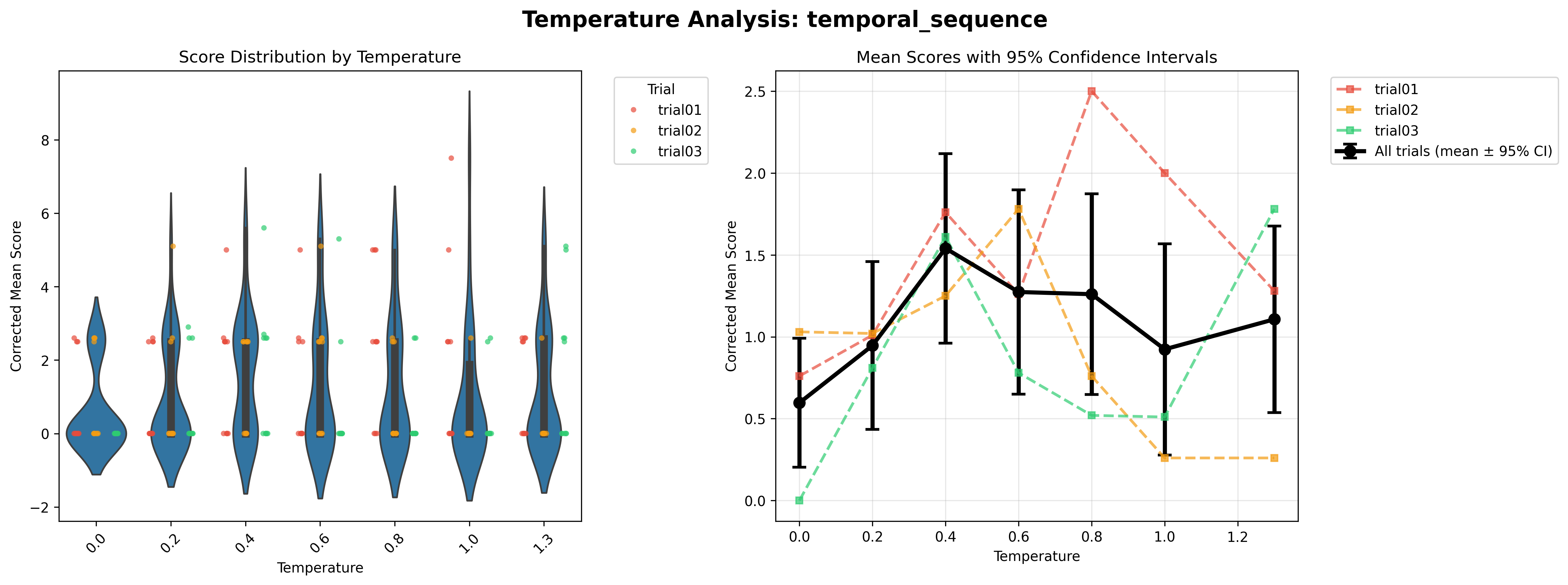}
\caption{Temperature parameter analysis for Temporal Sequences}
\label{fig:temp_temporal}
\end{figure}

\begin{figure}[htbp]
\centering
\includegraphics[width=0.9\textwidth]{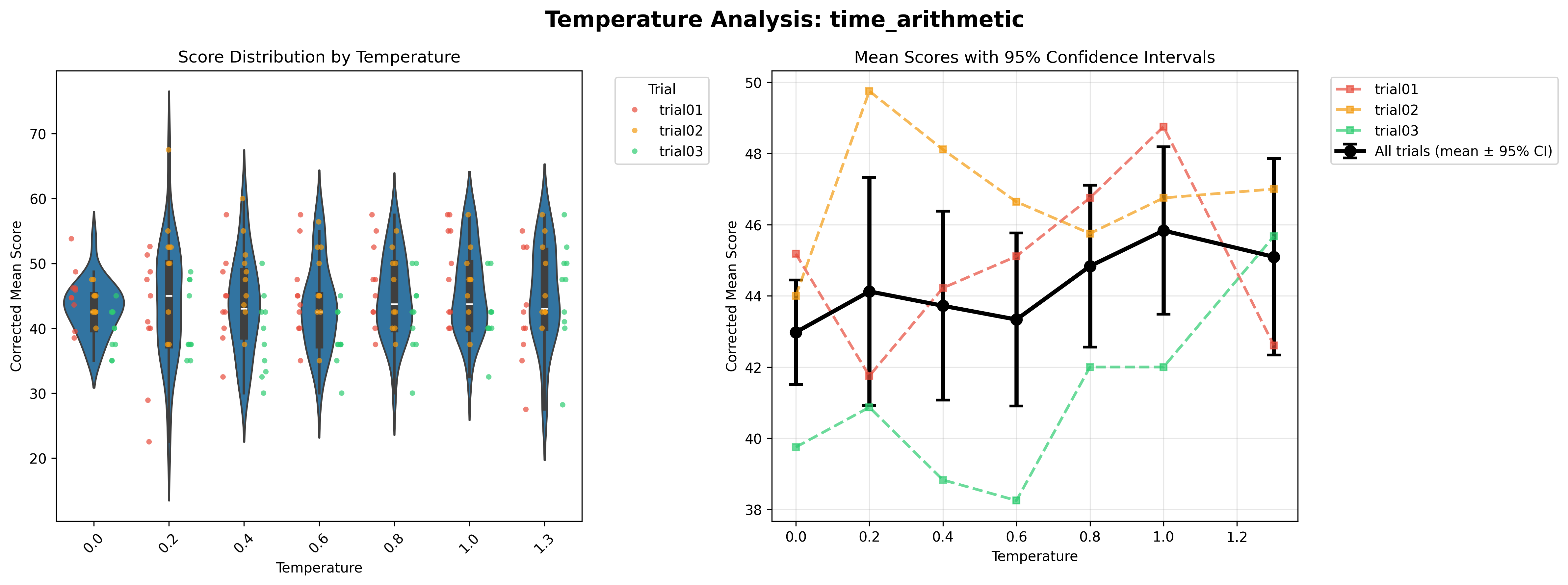}
\caption{Temperature parameter analysis for Time Arithmetic}
\label{fig:temp_time}
\end{figure}

\begin{figure}[htbp]
\centering
\includegraphics[width=0.9\textwidth]{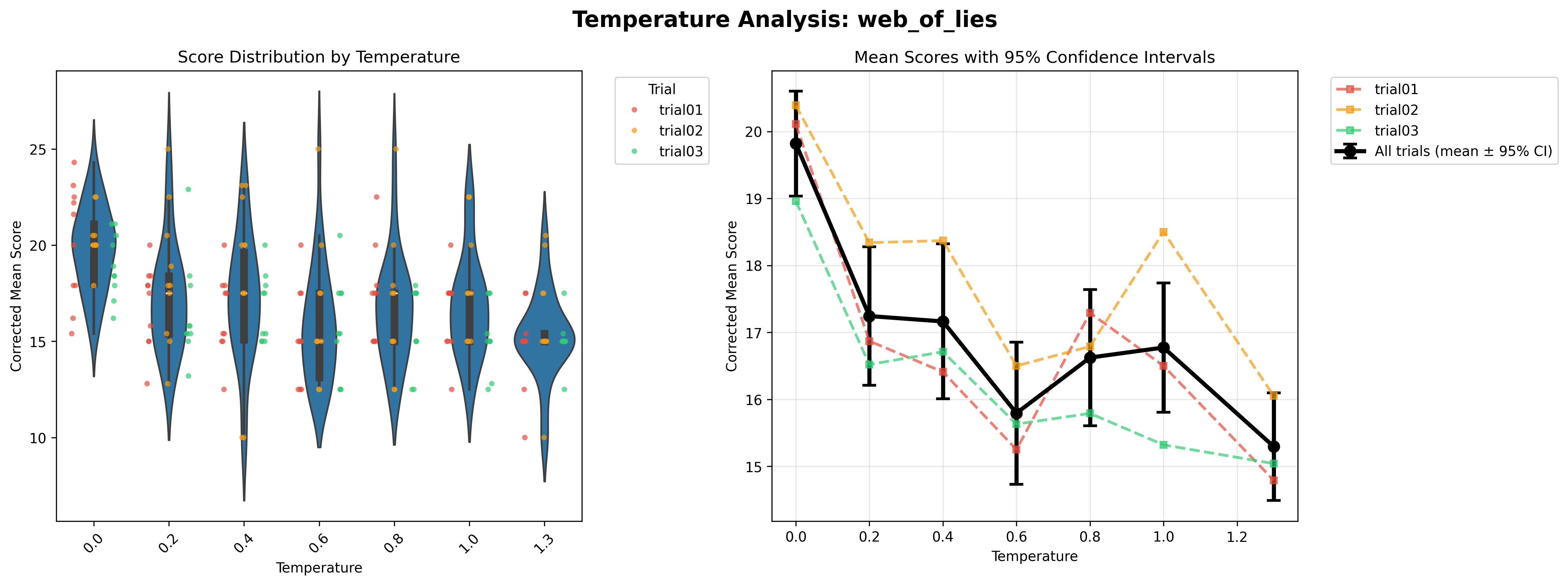}
\caption{Temperature parameter analysis for Web of Lies}
\label{fig:temp_web}
\end{figure}

\begin{figure}[htbp]
\centering
\includegraphics[width=0.9\textwidth]{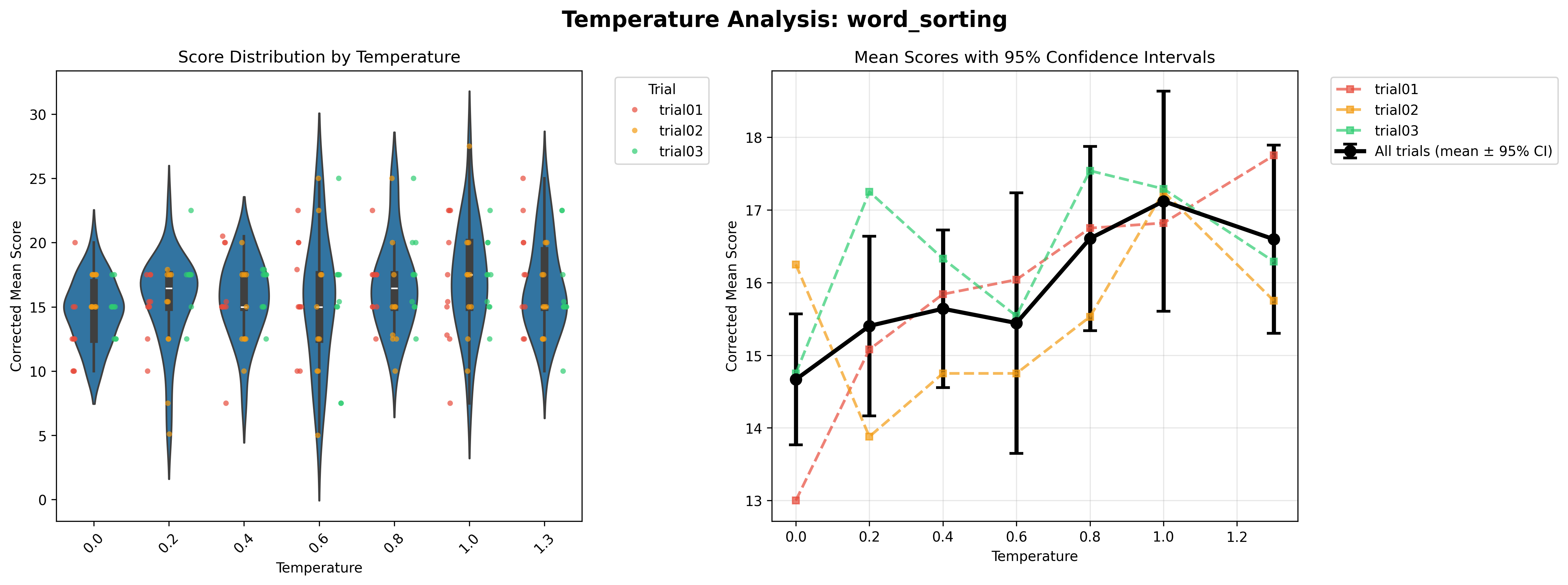}
\caption{Temperature parameter analysis for Word Sorting}
\label{fig:temp_word}
\end{figure}

\begin{figure}[htbp]
\centering
\includegraphics[width=0.9\textwidth]{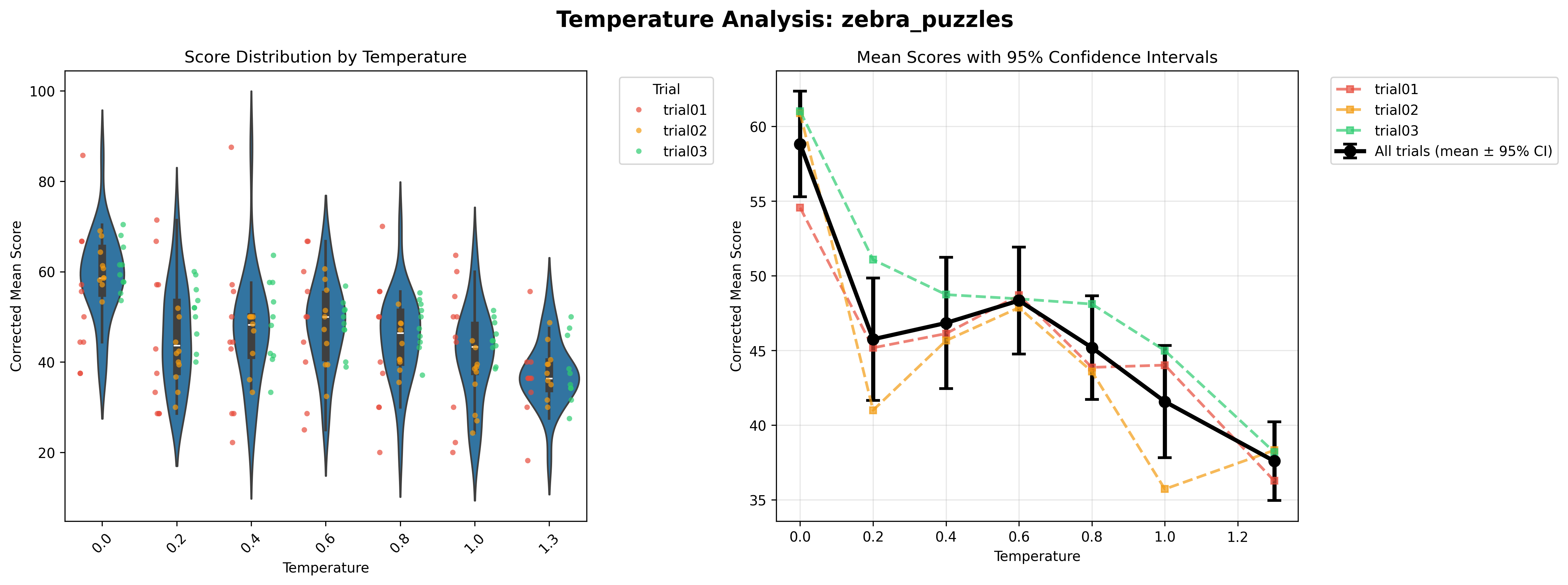}
\caption{Temperature parameter analysis for Zebra Puzzles}
\label{fig:temp_zebra}
\end{figure}

\FloatBarrier

\end{document}